\documentclass{article}
\usepackage[utf8]{inputenc}
\pdfoutput=1
\usepackage{hyperref}
\usepackage[english]{babel}
\usepackage{natbib}
\usepackage{graphics}
\usepackage{graphicx}
\usepackage{caption}
\usepackage{subcaption}
\usepackage{comment}
\usepackage{booktabs}
\usepackage[T1]{fontenc}
\usepackage{authblk}
\usepackage{orcidlink}

\usepackage[arrowdel]{physics}
\usepackage{blindtext}

\title{KASAM: Spline Additive Models for Function Approximation}
\author[]{Heinrich van Deventer \orcidlink{0000-0001-9309-4330} \thanks{HPDeventer@gmail.com}}
\author[]{Pieter Janse van Rensburg}
\author[]{Anna Bosman \orcidlink{0000-0003-3546-1467}
\thanks{anna.bosman@up.ac.za}}
\affil[]{Department of Computer Science, University of Pretoria}

\date{\today}

\begin{document}

\maketitle

\begin{abstract}
% The abstract is under the limit of 200 words
    Neural networks have been criticised for their inability to perform continual learning due to catastrophic forgetting and rapid unlearning of a past concept when a new concept is introduced. Catastrophic forgetting can be alleviated by specifically designed models and training techniques. This paper outlines a novel Spline Additive Model (SAM). SAM exhibits intrinsic memory retention with sufficient expressive power for many practical tasks, but is not a universal function approximator. SAM is extended with the Kolmogorov-Arnold representation theorem to a novel universal function approximator, called the Kolmogorov-Arnold Spline Additive Model - \textit{KASAM}. The memory retention, expressive power and limitations of SAM and KASAM are illustrated analytically and empirically. SAM exhibited robust but imperfect memory retention, with small regions of overlapping interference in sequential learning tasks. KASAM exhibited greater susceptibility to catastrophic forgetting. KASAM in combination with pseudo-rehearsal training techniques exhibited superior performance in regression tasks and memory retention.
    
\end{abstract}
 %If some of the constant parameters in KASAM are instead randomly initialised and trainable as with a conventional ANN, then its performance is terribly degraded. 
% keywords: catastrophic interference, catastrophic forgetting

%%%%%%%%%%%%%%%%%%%%%%%%%%%%%%%%%%%%%%%%%%%%%%%%%%%%%%%%%%%%%%%%%%%%%%%%%%%%%%%%%%%%%%%%%
\section{Introduction}
%%%%%%%%%%%%%%%%%%%%%%%%%%%%%%%%%%%%%%%%%%%%%%%%%%%%%%%%%%%%%%%%%%%%%%%%%%%%%%%%%%%%%%%%%

Continual learning refers to the ability of a model to learn from a stream of incoming data sequentially over time, while retaining knowledge acquired from previous data. Continual learning is a vital component of machine learning. It enables a model to generalise in situations where the stream of data may be non-stationary, with unavailable data during training, or when new information is incrementally made available to the model over time~\citep{kirkpatrick2017overcoming}.

A phenomenon called catastrophic forgetting hinders continual learning in many artificial neural networks (ANNs)~\citep{howard2018universal, schak2019study}. Catastrophic forgetting refers to the model losing knowledge of previous datasets or tasks as the model is trained sequentially on information relevant to a new dataset or task~\citep{kirkpatrick2017overcoming}. Catastrophic forgetting is also called catastrophic interference in older works~\citep{MCCLOSKEY1989109}. 

If a model, e.g. an ANN, has very robust memory that is not susceptible to catastrophic forgetting, then such a model may be susceptible to overfitting (i.e. memorisation of the training set). Overfitting leads to poor generalisation. Humans, however, have both reasonable (albeit imperfect) memory retention and good generalisation, so in principle it should be possible to build a model with similar desirable properties.

Splines are piece-wise defined functions. The application of splines to mitigate catastrophic forgetting was absent in most of the reviewed literature. Due to the piece-wise definition, each spline parameter only affects the function on some small region while keeping the rest of the function unchanged, thus making it a good candidate for continual learning. Cubic B-splines are considered in this paper.

Catastrophic forgetting is typically mitigated with two broad strategies: carefully designed and parameterised models, or through augmented training and regularisation techniques ~\citep{robins1995catastrophic,shin2017continual}. This paper attempts to address catastrophic forgetting through the following novel contributions:

\begin{itemize}
    \item A novel Spline Additive Model (SAM) with guaranteed robustness to catastrophic forgetting is proposed, which is useful for many applications. However, it is not a universal function approximator.
    \item The Kolmogorov-Arnold Spline Additive Model (KASAM) is proposed, which is a novel architecture that combines SAMs with the Kolmogorov-Arnold representation theorem to create a universal function approximator.
\end{itemize}

These goals demand reviewing the fundamentals of function approximation in one variable, with B-spline functions that are resistant to catastrophic forgetting. The paper proceeds to build multi-variable function approximators with single-variable B-spline functions and the Kolmogorov-Arnold representation theorem.

The rest of the paper is structured as follows: Section~\ref{sec:previous} provides an overview of the related literature. Section~\ref{sec:catastrophic} illustrates the concept of catastrophic forgetting. Section~\ref{sec:singleVar} lists the properties of splines in the context of single-variable function approximation. Section~\ref{sec:sam} discusses SAMs as function approximators. Section~\ref{sec:KASAM} introduces the KASAM architecture. Section~\ref{sec:pseudo} describes the pseudo-rehearsal technique employed in this study. Section~\ref{sec:method} details the methodology. Section~\ref{sec:exp} presents the empirical results. Section~\ref{sec:conclusions} concludes the paper, and Section~\ref{sec:opportunities} proposes some directions for future research.

%%%%%%%%%%%%%%%%%%%%%%%%%%%%%%%%%%%%%%%%%%%%%%%%%%%%%%%%%%%%%%%%%%%%%%%%%%%%%%%%%%%%%%%%%
\section{Relevant Studies}\label{sec:previous}
%%%%%%%%%%%%%%%%%%%%%%%%%%%%%%%%%%%%%%%%%%%%%%%%%%%%%%%%%%%%%%%%%%%%%%%%%%%%%%%%%%%%%%%%%

%Similarly, zero-centred weight regularisation (like ridge or lasso regression) can also affect the model's parameters, leading to catastrophic forgetting. Random perturbations and evolutionary training algorithms (depending on the specific implementation) can also affect the neural network's parameters during training unpredictably. 

It has been hypothesized that overlapping, dense and distributed representations in ANNs lead to catastrophic forgetting~\citep{kaushik2021understanding}. Catastrophic forgetting occurs when many parameter estimates that store knowledge for one task change during sequential learning to meet the objectives of another task~\citep{kirkpatrick2017overcoming, mcrae1993catastrophic}. If the same parameter is shared or overlaps with many inputs, then it is more susceptible to catastrophic forgetting. Any gradient-based updates would affect the same shared parameter and thus, the parameter value would be more likely to change between tasks. Catastrophic forgetting can be ameliorated with models that are parameterised in such a way that weight sharing or overlap is minimised over all inputs.

Training techniques to counteract catastrophic forgetting include identifying and protecting key parameters for different tasks. Parameter regularisation to penalise adjusting parameters from their initial values is one approach. Retraining over all training data for all tasks can also be done, although this scales poorly as the amount of training data increases. 

Data augmentation techniques can also be used to counteract catastrophic forgetting, some of which are referred to as rehearsal techniques. There are many suggested rehearsal and pseudo-rehearsal techniques~\citep{robins1995catastrophic}. Pseudo-rehearsal works quite well for low dimensional problems, but there is some experimental evidence and practical use cases that suggest pseudo-rehearsal has poorer performance in high dimensional problems. It is hypothesised that the concentration of measure in high dimensional probability distributions require more complexity to be modelled for acceptable results. As a consequence, some researchers considered estimating the data's distribution using a Generative Adversarial Network (GAN) for higher dimensional problems. %Generative replay involves using a generative model, like a GAN, to sample from the previously seen data's distribution and to augment the training data~\citep{shin2017continual}. 
However, training GANs requires a lot of computational resources, and might not be a scalable solution for all problems and use-cases.

Using splines in ANNs has been studied to some extent. \citet{scardapane2017learning} studied learnable activation functions parameterised by splines. \citet{scardapane2017learning} introduced vectorised gradient descent based methods to train the parameters of their architecture.  \citet{scardapane2017learning} tested the function approximation ability of the architecture. \citet{douzette2017b} made use of spline networks, allowing trainable and non-uniform sub-intervals, and developed algorithms for evaluating splines and their derivatives with respect to all their parameters. The research concluded that splines which allowed non-uniform partitions that vary during training achieved state-of-the-art results for spline-based architectures. The accuracy of non-uniform splines compared well against conventional neural networks. However, allowing the partitions of sub-intervals to vary had counter-productive effects. Intermediate or hidden layers could take on values outside the support interval of the splines. Since splines are zero outside their sub-intervals, this could lead to increased training times~\citep{douzette2017b}. To the best of the authors' knowledge, catastrophic forgetting was not considered in the context of splines.

A lot of research has taken place on the neural network approximations of the Kolmogorov-Arnold representation theorem~\citep{funahashi1989approximate,scarselli1998universal,igelnik2003kolmogorov,braun2009constructive,guliyev2018approximation,sannai2019universal, schmidt2021kolmogorov,shen2021neural}. \citet{shen2021neural} gives a good overview of the direction of research, and has shown that the focus has been on reducing the number and width of the hidden layers required to approximate the theorem with a neural network, at the trade-off of increasing the bounded approximation error. The difference between the well-known universal function approximation theorems for arbitrarily deep or arbitrarily wide neural network and the Kolmogorov-Arnold representation theorem is striking.

ANN architectures using splines and the Kolmogorov-Arnold representation theorem have also been explored~\citep{igelnik2003kolmogorov,lane1991multi}. \citet{igelnik2003kolmogorov} put forth a Kolmogorov Spline Network architecture, which was also based on the Kolmogorov-Arnold representation theorem. However, this differed from our work through the manner in which the weights are applied~\citep{igelnik2003kolmogorov}. Furthermore, no prototype was constructed, and no experimental work was performed.

The NeurIPS (originally NIPS) paper \citet{lane1991multi} explored a Kolmogorov based multi-layer perceptron architecture which employed B-splines in the nodes. \citet{lane1991multi} trained the architecture with gradient-descent based techniques and tested out the function approximation ability. \citet{lane1991multi} states that the architecture was able to approximate functions, and commented on the fast rate of convergence of the network’s parameters. Their work is most relevant to this paper. KASAM has a skip-connection to the output and can more easily represent functions that are sum-decomposable into a sum of single-variable functions. This paper focuses on memory retention and catastrophic forgetting, whereas the paper \citet{lane1991multi} does not consider catastrophic forgetting.

The research by Numenta and Jeffrey Hawkins is focused on neuroscience and reverse-engineering the computational mechanisms employed in the brain~\citep{htm_neocortex}. Their research is based on discrete and stochastic models like hierarchical temporal memory with update rules that appear otherworldly compared to the familiar differentiable models that can be trained with gradient descent algorithms. One aspect that is missing is a universal function approximation theorem proving the expressive power of their models. One component of their technology is Sparse Distributed Representations (SDRs) that encode real numbers with sparse vectors  \citep{Hinton1990DistributedR}. SDRs are similar to zeroth order B-splines, although the connection has not been explicitly shown in the reviewed literature.

The potential use of B-spline functions to mitigate catastrophic forgetting was not thoroughly investigated in any of the reviewed literature. The convergence rate and numerical stability of B-splines require more thorough analysis and empirical study.

%%%%%%%%%%%%%%%%%%%%%%%%%%%%%%%%%%%%%%%%%%%%%%%%%%%%%%%%%%%%%%%%%%%%%%%%%%%%%%%%%%%%%%%%%
\section{Susceptibility to Catastrophic Forgetting}\label{sec:catastrophic}
%%%%%%%%%%%%%%%%%%%%%%%%%%%%%%%%%%%%%%%%%%%%%%%%%%%%%%%%%%%%%%%%%%%%%%%%%%%%%%%%%%%%%%%%%

Model parameters that are shared over the entire input-domain of a function approximator make models such as ANNs or linear functions susceptible to catastrophic forgetting. A parameter that affects a model over all inputs is a globally shared parameter, and not a localised parameter. Parameters that are localised only affect a model's output over a small region of the input-domain. Catastrophic forgetting is easily demonstrated with a simple linear function approximator for a single-variable scalar function. 

\begin{figure}[h!]
\centering
\noindent
\includegraphics[width=0.7\textwidth]{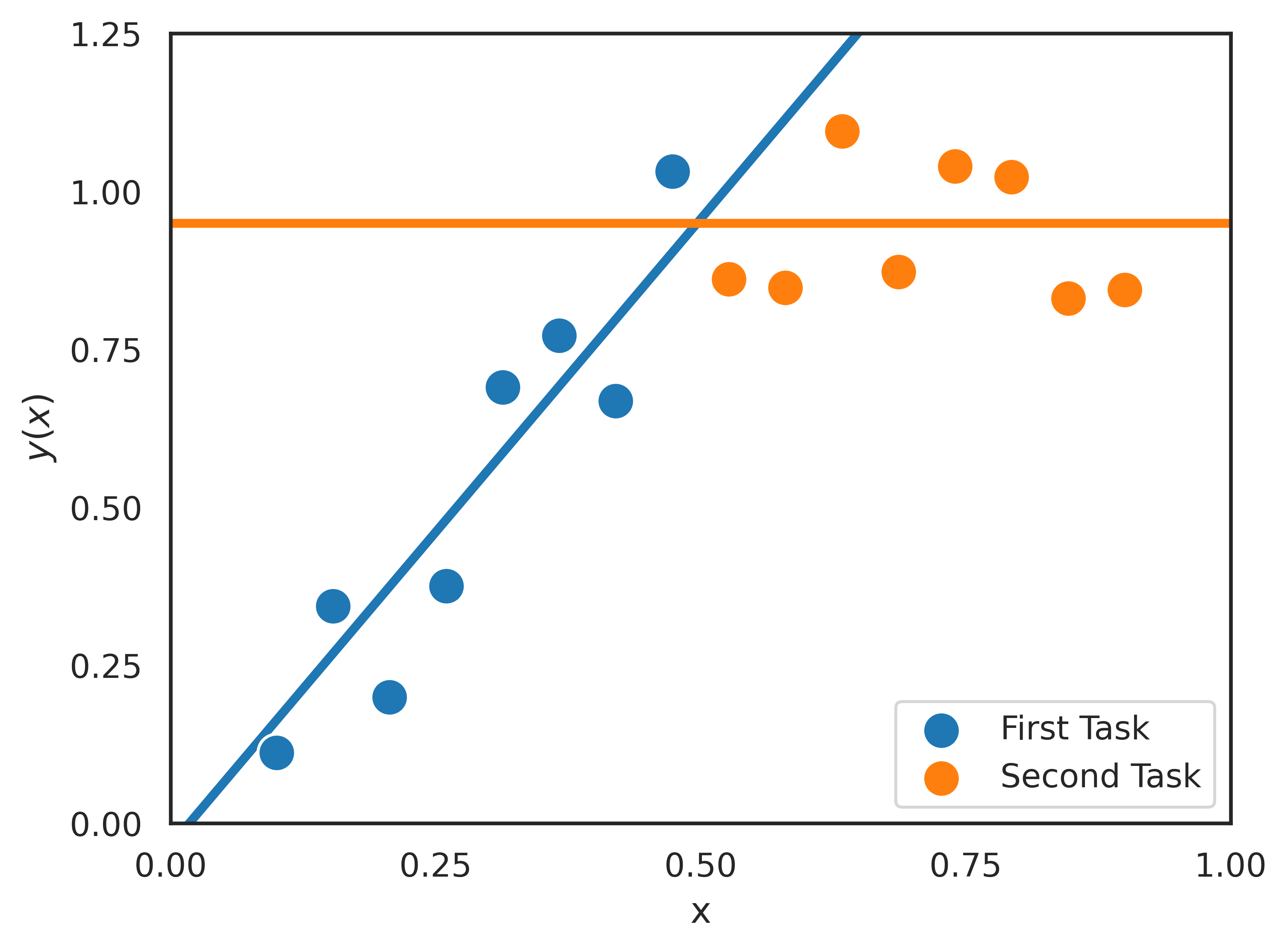}
\caption{\small A linear model with only two parameters is trained on data for the first task. The same model is then trained on data from a second task, and will adapt to the new task. The model abruptly forgets about the first task without revision. Trainable parameters that are globally shared across the input-domain make the model susceptible to catastrophic forgetting.}
\label{fig:fig_label_linear_functions}
\end{figure}

A linear model is trained on the data sampled from the distribution for the first task in Figure~\ref{fig:fig_label_linear_functions}. The model is thereafter trained on the second task. Without any additional guidance, such as revision of the first task or weight regularisation to prevent catastrophic forgetting, the model promptly unlearns the first task.

If the linear model had been trained on both tasks simultaneously, then it would have found a reasonable solution for both tasks. The individual tasks have different data distributions, which affects the out-of-sample performance and the potential for continual or incremental learning. The joint distribution for both tasks has a non-linear optimal function (which is piece-wise linear for this specific example). 

The seemingly disparate ideas of catastrophic forgetting, distribution shift, out-of-sample performance, continual learning, and the non-linearity of a target function are related. The relationship is not a simple analytical rule, but it is worth investigating.

%%%%%%%%%%%%%%%%%%%%%%%%%%%%%%%%%%%%%%%%%%%%%%%%%%%%%%%%%%%%%%%%%%%%%%%%%%%%%%%%%%%%%%%%%
\section{Splines for Single Variable Function Approximation}\label{sec:singleVar}
%%%%%%%%%%%%%%%%%%%%%%%%%%%%%%%%%%%%%%%%%%%%%%%%%%%%%%%%%%%%%%%%%%%%%%%%%%%%%%%%%%%%%%%%%
% can use any set of basis functions, it is recommended to use basis functions that are orthonormal,  % with gradient flow attenuation % uniform cubic b-spline, number of basis elements is fixed number. Show stratification % uniform cubic b-splines with sufficiently many knots have gradients updates that are zero almost everywhere, and the gradient vector is orthogonal to the gradient vector of another point that is sufficiently far away; insert the rigorous definition of the cross-shaped neighborhood that is considered in this case.

\begin{figure}[h]
\centering
\noindent
\includegraphics[width=0.7\textwidth]{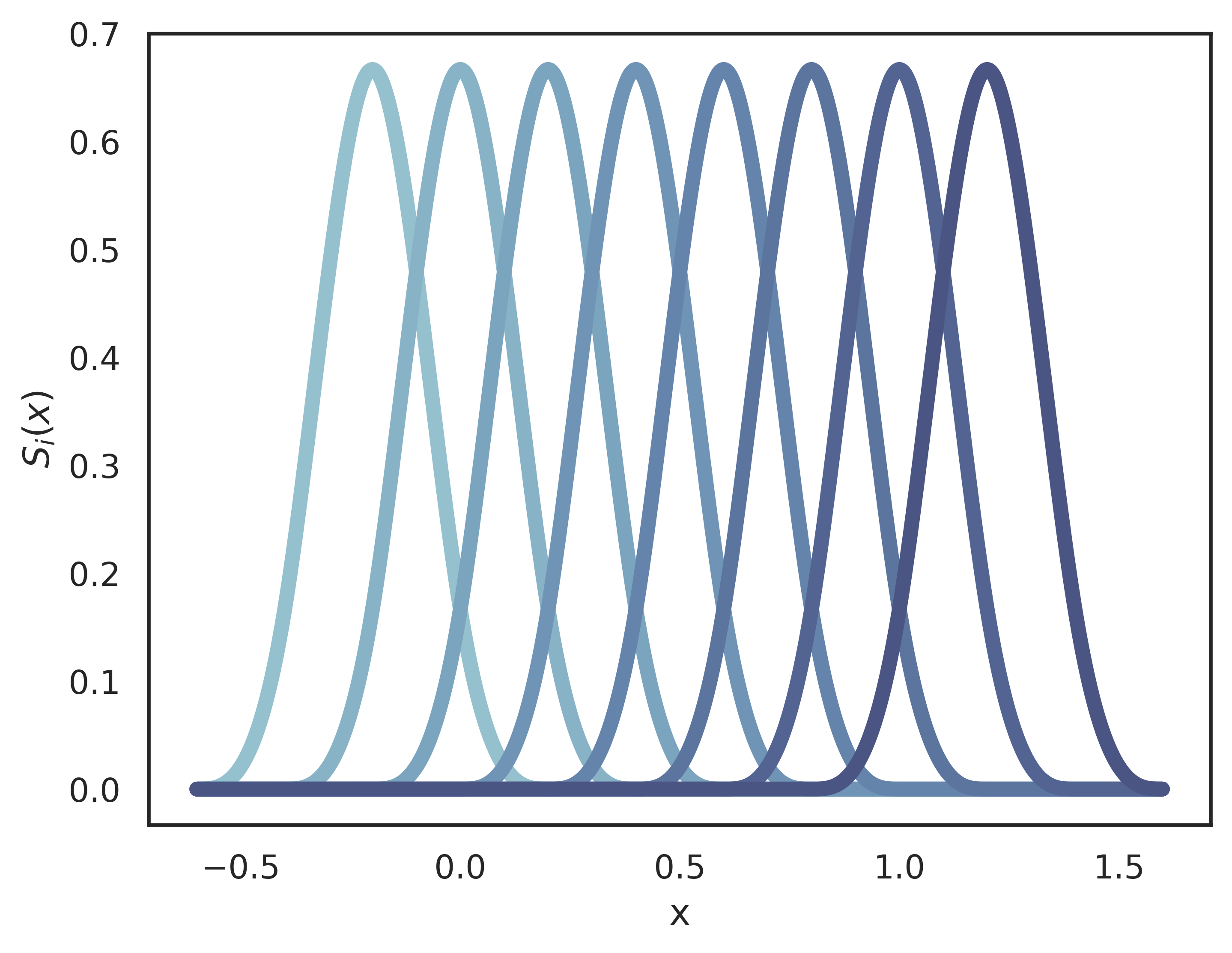}
\caption{\small Uniform cubic B-spline basis functions with a density of 8. Note that the basis functions have been plotted for input values outside the unit interval to clearly see all 8 basis functions. Each basis function is zero almost everywhere except on a small sub-interval. Each basis function is a scaled and translated version of the same function such that $ S_{i}(x) = S(w_{i}x + b_{i})$. }
\label{fig:fig_label_uniform_cubic_spline_basis_functions}
\end{figure}

Single-variable function approximation on the unit interval is typically done with the use of single-variable basis functions. Examples include the polynomial basis (Legendre polynomials), or the Fourier basis. Using a large density (i.e. number) of basis functions can lead to high-variance models with extreme oscillations between the finitely many training data points. It is an open question if there are regularisation methods that attenuate such oscillations for any choice of basis functions. The specific basis functions considered in this paper are uniform cubic B-splines, illustrated in Figure~\ref{fig:fig_label_uniform_cubic_spline_basis_functions}.

The use of B-splines is a flexible and expressive method of function approximation. The analysis of B-splines and characterising their general behaviour is simple compared to other basis functions. Note that the number of basis functions is fixed and chosen independently of the training data. The model considered in this paper does not necessarily interpolate between consecutive data points. Exact interpolation of training data is ill-advised for tasks with noisy training data, since the resulting model would have a lot of variance and oscillate wildly. The order of a B-spline (e.g. zeroth or cubic) is different to the number of basis functions. The number or density of basis functions is sometimes referred to as sub-intervals or knots in the literature on splines and B-splines.
%It should be noted that when u1sed for regression or classification tasks, the number of basis functions fixed and chosen independently of the number of training data points. In other words, the model does not interpolate between each consecutive training data point on the unit interval. The order of a B-spline (zeroth, cubic etc.) is different to the number of knots, sub-intervals or basis functions.

%%%%%%%%%%%%%%%%%%%%%%%%%%%%%%%%%%%%%%%%%%%%%%%%%%%%%%%%%%%%%%%%%%%%%%%%%%%%%%%%%%%%%%%%%
% activation function definition 
%%%%%%%%%%%%%%%%%%%%%%%%%%%%%%%%%%%%%%%%%%%%%%%%%%%%%%%%%%%%%%%%%%%%%%%%%%%%%%%%%%%%%%%%%

Each basis function for a uniform cubic B-spline $S_{i}$ can be obtained by scaling and translating the input of the following activation function:
$$ S(x) =\begin{cases} 
      \frac{1}{6} x^{3} &  0 \leq x < 1\\
      \frac{1}{6} \left[-3(x-1)^{3} +3(x-1)^{2} +3(x-1) + 1 \right] &  1 \leq x < 2\\
      \frac{1}{6} \left[3(x-2)^{3} -6(x-2)^{2} + 4 \right]  & 2 \leq x < 3\\
      \frac{1}{6} ( 4-x ) ^{3} &  3 \leq x < 4\\
      0 & otherwise 
   \end{cases}
$$

%%%%%%%%%%%%%%%%%%%%%%%%%%%%%%%%%%%%%%%%%%%%%%%%%%%%%%%%%%%%%%%%%%%%%%%%%%%%%%%%%%%%%%%%%

\begin{figure}[!h]
\centering
\noindent
\includegraphics[width=0.8\textwidth]{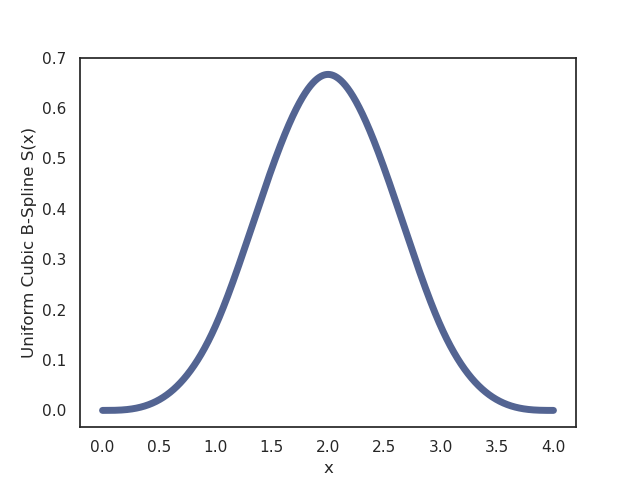}
\caption{\small The activation function used to implement uniformly spaced cubic B-splines. In the uniform case all basis functions are the same shape with only the input or argument being scaled and translated with a fixed linear function for each basis function.}
\label{fig:fig_label_uniform_cubic_spline_activation}
\end{figure}

% 
%%%%%%%%%%%%%%%%%%%%%%%%%%%%%%%%%%%%%%%%%%%%%%%%%%%%%%%%%%%%%%%%%%%%%%%%%%%%%%%%%%%%%%%%%
The initial impetus for considering B-splines followed from the revelation that Numenta's Sparse Distributed Representations (SDRs) \citep{Hinton1990DistributedR} are mathematically similar to zeroth-order B-splines. Zeroth order splines incidentally resemble lookup tables, and are not differentiable over their entire domain. It is also known that lookup tables are very robust to catastrophic forgetting \citep{look_up_tables_are_robust}. Replacing zeroth-order B-splines with cubic B-splines yields a differentiable model that is smooth and amicable to gradient descent-based learning methods. If the unit interval is uniformly partitioned B-splines as in Figure~\ref{fig:fig_label_uniform_cubic_spline_basis_functions}, then the subsequent set of basis functions have the same shape, as illustrated in Figure~\ref{fig:fig_label_uniform_cubic_spline_activation}.

%The modern machine learning libraries used for creating and training artificial neural networks are not typically used for single-variable function approximation using basis functions. Finding a way to implement such a function approximation scheme required out-of-the-box thinking.

Each cubic B-spline basis function is simply a re-scaled or translated version of any other cubic B-spline basis function, as seen in Figures~\ref{fig:fig_label_uniform_cubic_spline_basis_functions} and~\ref{fig:fig_label_uniform_cubic_spline_activation}. This symmetry can promote reuse. In a neural network context, uniform cubic B-splines can be modelled using a two-layer neural network with an activation function corresponding to the shape of each basis function, and a pre-defined set of untrainable weights and biases. A final trainable linear layer multiplies each basis function with its coefficient parameter and sums together the results. A single variable function approximator for uniform B-splines is given by:

$$
f(x) 
= \sum_{i=1}^{K} \theta_{i} S_{i}(x) 
= \sum_{i=1}^{K} \theta_{i} S(w_{i}x + b_{i})
$$

%Cubic B-splines have three properties that are atypical of most function approximators, namely sparsity, bounded gradients, and orthogonal distal gradients. These properties hold for arbitrarily many basis functions. The listed properties are specified below for cubic B-splines, but similar properties hold for any fixed order B-spline functions. 

This construction presented above is similar to other basis functions. The Fourier series is composed of basis functions that are orthogonal to each other. There are many analytical and practical reasons for considering basis functions that are orthogonal with respect to each other. The Fourier basis for functions on the unit interval is:

$$
f(x) 
= c + \sum_{n=1}^{K} a_{n} \sin(2 \pi n x) + b_{n} \cos(2 \pi n x)
$$

Polynomial functions can also constitute a set of basis functions. The relatively simple monomial basis is non-orthogonal and of the form:

$$
f(x) 
= c + \sum_{n=1}^{K} a_{n} x^{n}
$$

In many applications it is preferable to use orthogonal basis functions. Legendre polynomials are orthogonal polynomial basis functions. A function approximator in terms of Legendre basis functions $P_{n}(x)$:

$$
f(x) 
=  \sum_{n=0}^{K} a_{n} P_{n}(x)
$$

The choice of B-splines seems arbitrary, but it is a critical design choice. The problem with trigonometric and polynomial functions is that they are non-zero almost everywhere. There are only finitely many points where each basis function is zero. This means that each parameter could affect a model's output almost everywhere and lead to catastrophic forgetting. B-splines are zero almost everywhere and don't suffer such off-target effects - each parameter affects only a small region. Cubic B-splines have three properties that are atypical of most function approximators, namely sparsity, bounded gradients, and orthogonal distal gradients. These properties hold for arbitrarily many basis functions. The listed properties are specified below for cubic B-splines, but similar properties hold for any fixed order B-spline functions.

%%%%%%%%%%%%%%%%%%%%%%%%%%%%%%%%%%%%%%%%%%%%%%%%%%%%%%%%%%%%%%%%%%%%%%%%%%%%%%%%%%%%%%%%%
\subsection*{Property 1: Sparsity of the Gradient Vector}
%%%%%%%%%%%%%%%%%%%%%%%%%%%%%%%%%%%%%%%%%%%%%%%%%%%%%%%%%%%%%%%%%%%%%%%%%%%%%%%%%%%%%%%%%

%\textit{Property 1: Sparsity of the gradient vector} \\
%\textit{Property 1: } \\
The maximum number of non-zero cubic B-spline basis functions for any input is four, regardless of the number of basis functions $K$:

$$ 
\norm{ \grad_{\vec{\mathbf{\theta}}} f(x)}_{0} 
:=  \sum_{i=1}^{K}  d_{Hamming} \left(\frac{\partial f}{\partial \theta_{i}}(x),0  \right) 
\leq 4
\; \forall x \in D(f) 
$$

Sparsity is related to the robustness of a model to catastrophic forgetting. Uniform cubic B-splines require a minimum of four  basis functions over the unit interval, so $K\geq 4$. For very large models with a large density of basis functions the gradient vector is zero for nearly all trainable parameter as shown in Figure~\ref{fig:Proof_properties_1_2}.

%%%%%%%%%%%%%%%%%%%%%%%%%%%%%%%%%%%%%%%%%%%%%%%%%%%%%%%%%%%%%%%%%%%%%%%%%%%%%%%%%%%%%%%%%
\subsection*{Property 2: Gradient Flow Attenuation}
%%%%%%%%%%%%%%%%%%%%%%%%%%%%%%%%%%%%%%%%%%%%%%%%%%%%%%%%%%%%%%%%%%%%%%%%%%%%%%%%%%%%%%%%%

%\textit{Property 2: Gradient flow attenuation} \\
%\textit{Property 2: } \\
The gradient vector has a bounded L1 norm, for any number of B-spline basis functions:

$$ 
\norm{ \grad_{\vec{\mathbf{\theta}}} f(x)}_{1} 
= \sum_{i=1}^{K}  \left| \frac{\partial f}{\partial \theta_{i}} (x) \right| 
< 4U
\; \forall x \in D(f)
$$

\begin{figure}[!ht]
\centering
\noindent
\includegraphics[width=0.6\textwidth]{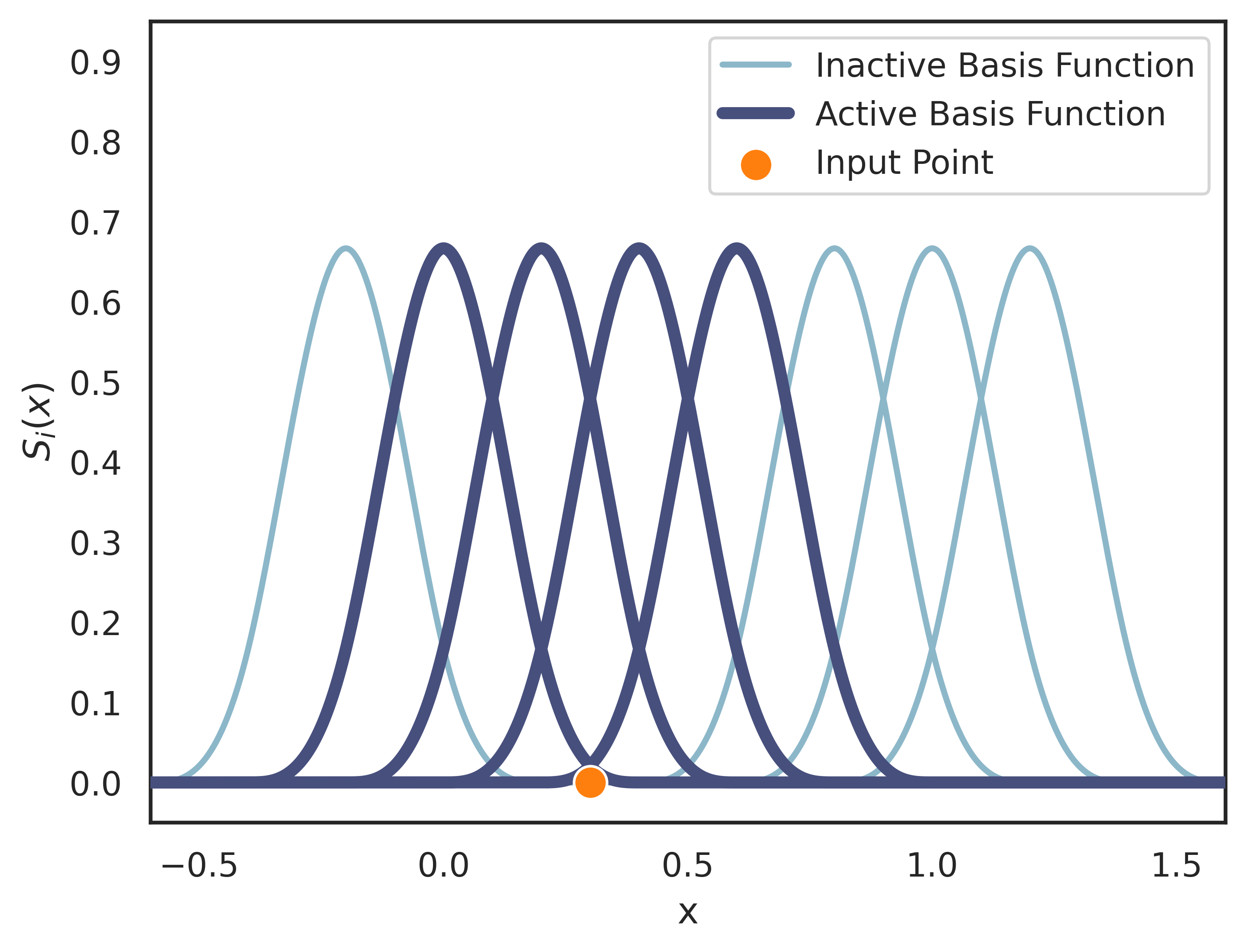}
\caption{\small Plot showing which basis functions are active for a single input. At most four basis functions are non-zero for any given input to a uniform cubic B-spline function, so one has the property of sparsity. Each of the four active basis functions that are active are bounded, so the sum of their absolute values is also bounded. Thus the $L_{1}$ norm of the gradient vector with respect to the trainable parameters is bounded.}
\label{fig:Proof_properties_1_2}
\end{figure}

 Gradient flow attenuation affects the stability of training a model with many trainable parameters. If the gradient vector is bounded, then the flow of gradient updates during training is attenuated and numerically stable. The visual proof of the boundedness of the gradient vector is demonstrated with Figure~\ref{fig:Proof_properties_1_2}.

%%%%%%%%%%%%%%%%%%%%%%%%%%%%%%%%%%%%%%%%%%%%%%%%%%%%%%%%%%%%%%%%%%%%%%%%%%%%%%%%%%%%%%%%%
%\subsection*{Distal Orthogonality}
\subsection*{Property 3: Distal Orthogonality}
%\subsubsection*{Property 3: Distal orthogonality}
%\subsection*{\textit{Property 3: Distal orthogonality}}
%\subsubsection*{\textit{Property 3: Distal orthogonality}}
%%%%%%%%%%%%%%%%%%%%%%%%%%%%%%%%%%%%%%%%%%%%%%%%%%%%%%%%%%%%%%%%%%%%%%%%%%%%%%%%%%%%%%%%%
%\textit{Property 3: Distal orthogonality} \\
%\textit{Property 3: } \\
For any uniform cubic B-spline function approximator of one variable there exists a $\delta>0$ such that: 

$$
|x - y| > \delta 
\implies  \langle \grad_{\vec{\mathbf{\theta}}} f(x),\grad_{\vec{\mathbf{\theta}}} f(y) \rangle = 0 
\; \forall x,y \in D(f)
$$

\begin{figure}[!h]
\centering
\noindent
\includegraphics[width=0.7\textwidth]{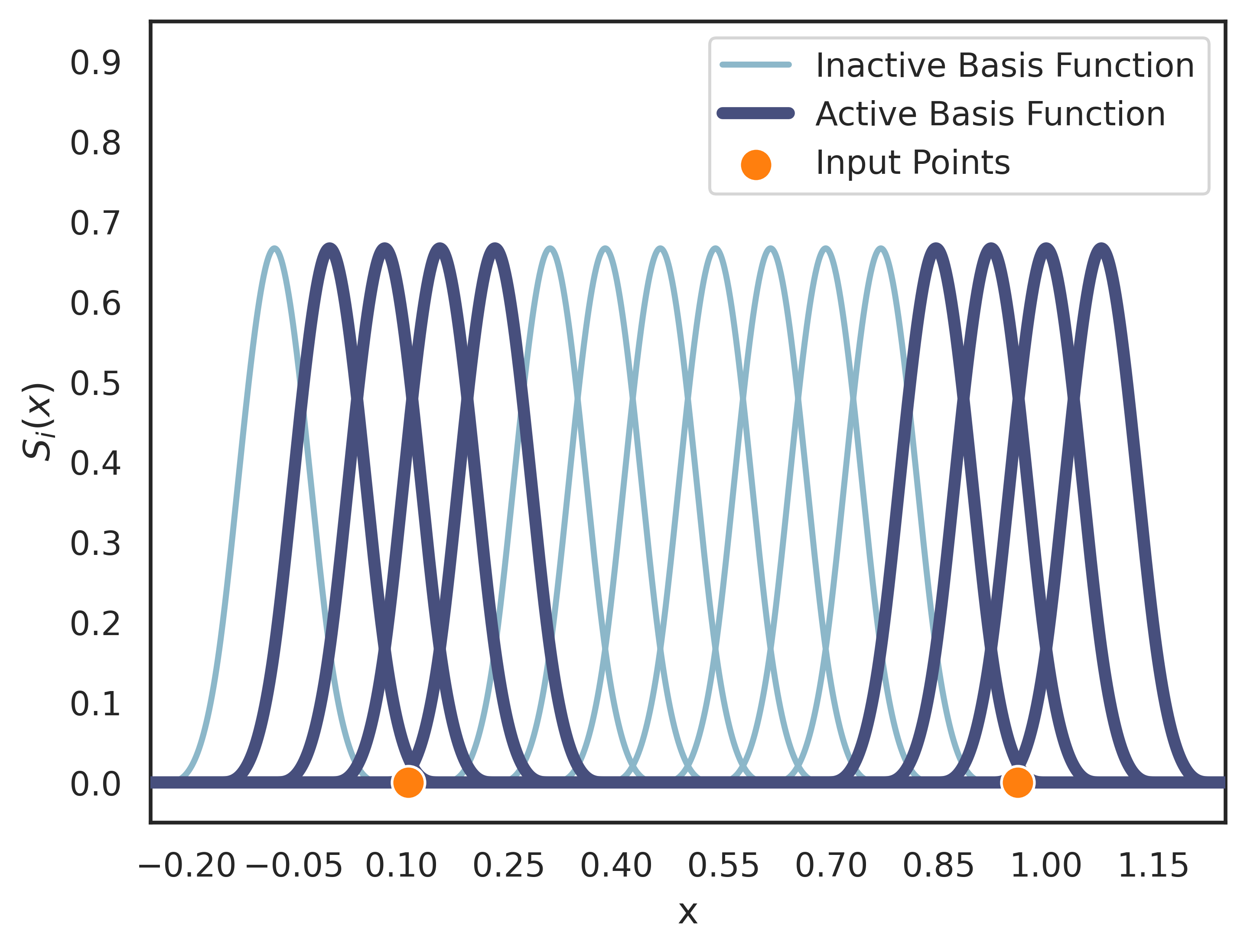}
\caption{\small Plot showing which basis functions are active for two distant input-points. 
If two input-points are sufficiently distant from each other, 
then there are no overlapping basis functions. 
The gradients with respect to trainable parameters are zero almost everywhere, 
except for the active basis functions. Thus, the inner-product of such gradient vectors must be zero. This property is best described as distal orthogonality.}
\label{fig:Proof_properties_3}
\end{figure}

If two input points are sufficiently far apart, then the gradient vectors with respect to the trainable parameters at each input are orthogonal to each other. Thus, only points within the same neighbourhood have non-orthogonal gradient vectors that can influence one another. A visual proof of distal orthogonality can be done with a diagram given in Figure~\ref{fig:Proof_properties_3}.

Distal orthogonality guarantees memory retention for single-variable function approximators based on cubic B-splines. Gradient flow attenuation guarantees bounded gradient vectors, which mean cubic B-splines are numerically stable during training, making optimisation much easier and well-posed. Sparsity of the gradient vector and sparse activation mean it is possible to implement very efficient models and training procedures that only compute the non-zero values. The above properties lead to a single-variable function approximator that is efficient, easily trained, and robust to catastrophic forgetting with intrinsic memory retention - in theory.

To the authors' knowledge, there are not many such guarantees given to other function approximators.

%$$|x - y| > \delta \implies  \nabla f(x) \cdot \nabla f(y) = 0 \; \forall x,y \in D(f) $$

%$$ |x - y| > \delta \implies  \grad_{\vec{\mathbf{\theta}}} f(x) \cdot \grad_{\vec{\mathbf{\theta}}} f(y) = 0 \; \forall x,y \in D(f) $$

%Other basis functions, such as Gaussian basis functions, were considered, but their use is ill-advised: While Gaussian functions do tend to zero, they are still non-zero everywhere over the input domain, which leads to susceptibility to catastrophic forgetting.

%%%%%%%%%%%%%%%%%%%%%%%%%%%%%%%%%%%%%%%%%%%%%%%%%%%%%%%%%%%%%%%%%%%%%%%%%%%%%%%%%%%%%%%%%
\subsection*{Stratification}
%%%%%%%%%%%%%%%%%%%%%%%%%%%%%%%%%%%%%%%%%%%%%%%%%%%%%%%%%%%%%%%%%%%%%%%%%%%%%%%%%%%%%%%%%

\begin{figure}[h]
\centering
\noindent
\includegraphics[width=0.6\textwidth]{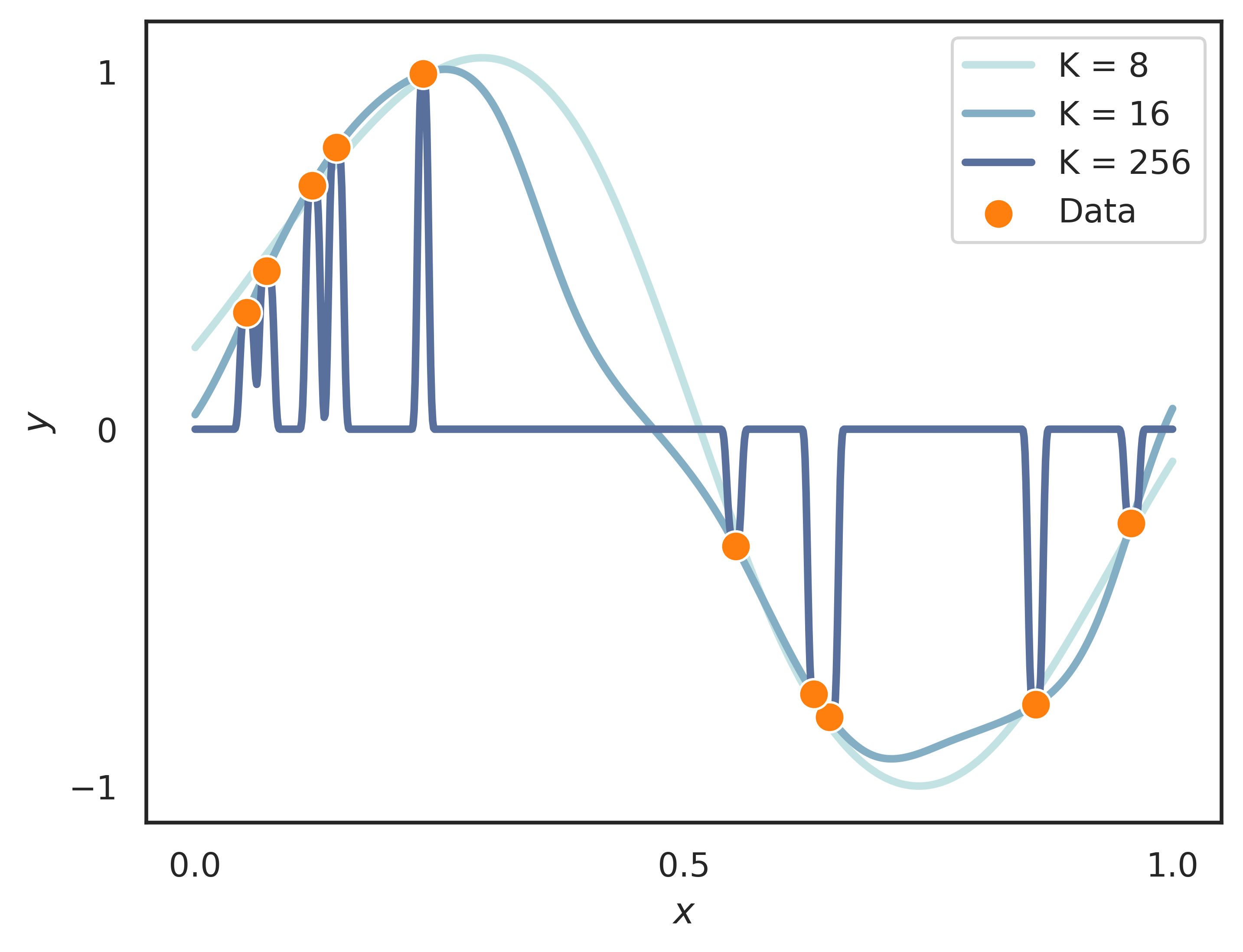}
\caption{\small Visualising stratification. The training data was sampled from the sine function $\sin(2 \pi x)$ to illustrate the effect of increasing the number basis function (denoted K). The uniform cubic B-spline models were initialised to be zero everywhere prior to training.}
\label{fig:fig_label_uniform_cubic_spline_stratification}
\end{figure}

Uniform cubic B-splines with a large density of basis functions can exhibit what is best described as stratification, illustrated in Figure~\ref{fig:fig_label_uniform_cubic_spline_stratification}. Stratification can be considered to be a special form of overfitting, where the data points are learned exactly, and the model is not adjusted for any regions not explicitly represented by the training data. The models in Figure~\ref{fig:fig_label_uniform_cubic_spline_stratification} are initialised to zero. Regions with no training data are never modified and retain their initial values. The gaps between data-points can thus lead to stratification, since only the regions with training data are updated. Therefore, over-parameterised B-spline models can easily memorise all training data.

% Stratification is a consequence of intrinsic memory retention. The flat unaltered regions between data-points are atypical, when contrasted with other basis functions.

There are a few qualitative differences between stratification and overfitting. Regions with no training data have predictable values and do not exhibit oscillations as seen in the Runge phenomenon with other kinds of basis functions~\cite{RungeBOYD199257,RungeFORNBERG2007379,RungeDEMARCHI2020112347}. Over-fitting is task-specific: Suppose the target function was known to be zero almost everywhere except at finitely many points, then an over-parametrised model would perform well. Anomaly detection is an ideal application of such over-parametrised B-spline models. On the other hand, over-parametrised B-spline models would perform poorly on regression problems like estimating a sine function with few training data-points as seen in Figure~\ref{fig:fig_label_uniform_cubic_spline_stratification}.

Stratification could be detrimental or advantageous, depending on the application. Manifold Mixup regularisation and data augmentation could be ideal strategies for counteracting stratification when necessary ~\cite{mixupBeyondEmpiricalRiskMinimization, ManifoldMixupBetterRepresentations,  MixUpasLocallyLinearOut-Of-ManifoldRegularization}.

%This occurs due to overfitting, as can be seen per Figure~\ref{fig:fig_label_uniform_cubic_spline_stratification}, where the over-parameterised uniform cubic B-spline perfectly fits the training data points, while retaining its initial value (which was chosen as 0 in this case) everywhere else. However, such an over-parameterised single variable uniform B-spline function approximator would not necessarily generalise well on some regression problems due to the gaps between finitely many training data points. %Some visualisations of this phenomenon resemble shattered glass or stratified rock-formations, hence the name. 
%

%In the next section, this approach is extended to multi-variable function approximation using Spline Additive Models. 
%There are numerous possibilities for extending this approach to multi-variable function approximation. One possibility is to use space-filling curves (e.g. fractals like Hilbert curves), or parameterised curves to map an input vector of many dimensions to the unit interval, and then apply single-variable function approximation. This could work well for some problems, but might be infeasible for problems in general. Using Spline Additive Models and the Kolmogorov-Arnold representation theorem might be more tractable.

% 
%%%%%%%%%%%%%%%%%%%%%%%%%%%%%%%%%%%%%%%%%%%%%%%%%%%%%%%%%%%%%%%%%%%%%%%%%%%%%%%%%%%%%%%%%
\section{Spline Additive Model (SAM)}\label{sec:sam}
%%%%%%%%%%%%%%%%%%%%%%%%%%%%%%%%%%%%%%%%%%%%%%%%%%%%%%%%%%%%%%%%%%%%%%%%%%%%%%%%%%%%%%%%%

% uniform cubic b-spline, show memory retention, 
%General Additive Models (GAMs) assume the target multi-variable function is the sum of single variable functions in each of the input variables. With $g_{j}$ being an approximation of $G_{j}$ which is an exact function. It is worth mentioning that linear functions are subsets of GAMs. General Additive Models (GAMs) are not general enough to approximate any multi-variable function, only some functions - this might be sufficient for many practical uses. GAMs in their original form, with $g_{j}$ being a single-variable function:

\begin{figure} [!h]
     \centering
     \begin{subfigure}[b]{0.4\textwidth}
         \centering
         \includegraphics[width=\textwidth]{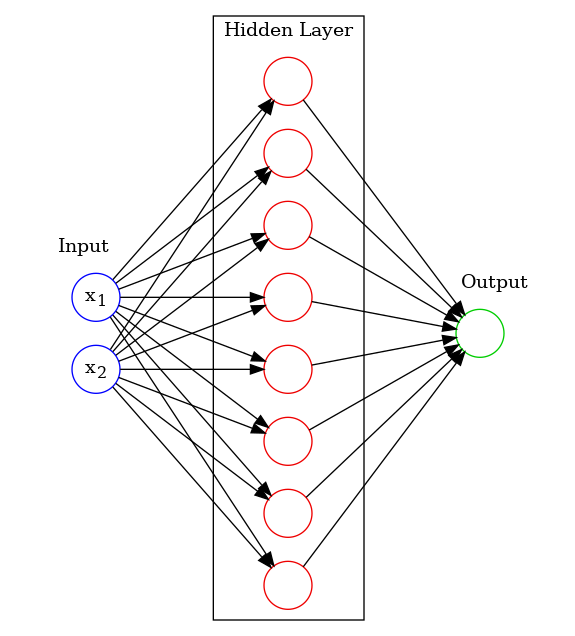}
         \caption{SAM with all weights}
         \label{fig:SAM_dense}
     \end{subfigure}
     \hfill
     \begin{subfigure}[b]{0.4\textwidth}
         \centering
         \includegraphics[width=\textwidth]{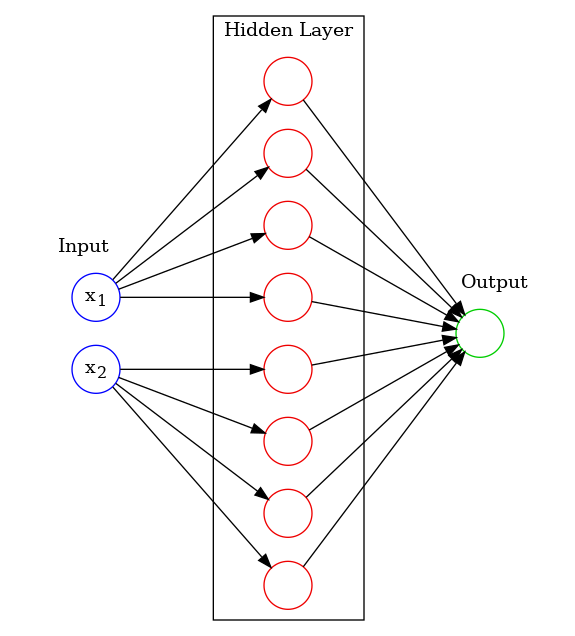}
         \caption{Non-zero or trainable weights}
         \label{fig:sam_dense}
     \end{subfigure}
        \caption{\small 
Structure of SAM. The weights and biases for the first layer are constant and specified with model creation to correctly implement B-splines.  The nodes in the black rectangle apply a non-linear activation function given by the cubic B-spline activation function. The output layer is a trainable linear layer with no bias term. The resulting model is a sum of single-variable functions in each parameter.}
        \label{fig:SAM_implementation}
\end{figure}

Extending single-variable function approximators to multi-variable function approximation is non-trivial. One possibility is to map a higher dimensional input to the unit interval using space-filling curves (e.g. fractals like Hilbert curves), or some other projection technique based on path integrals. In this paper, the choice was made to create a function approximator that is a sum of single variable functions of each input variable, called the Spline Additive Model (SAM) and illustrated in Figure~\ref{fig:SAM_implementation}.

Consider any target function $y$ that can be expressed as the sum of continuous single variable functions $y_{j}$ in each of the input variables $x_{j}$ given by: 

$$
y(\vec{\mathbf{x}}) 
= y(x_{1},...,x_{n}) 
= \sum^{n}_{j=1} y_{j}( x_{j} )
$$

SAM uses a uniform cubic B-spline function approximator $f_{j}$ with $K$ trainable parameters to approximate each single-variable function $y_{j}$. There are $K$ basis functions for each $f_{j}$, and there are $n$ input variables. The total number of trainable parameters is $nK$ for the entire model $f(\vec{\mathbf{x}})$. SAM inherits some of the properties discussed in Section~\ref{sec:singleVar}. SAM is given by the sum of $n$ single-variable B-spline functions:

$$
f(\vec{\mathbf{x}})  
= \sum^{n}_{j=1} f_{j}( x_{j} ) 
= \sum^{n}_{j=1} \sum_{i=1}^{K}   \theta_{i,j} S_{i,j}(  x_{j} )
= \sum^{n}_{j=1} \sum_{i=1}^{K}   \theta_{i,j} S(w_{i,j}x_{j} + b_{i,j})
$$

%%%%%%%%%%%%%%%%%%%%%%%%%%%%%%%%%%%%%%%%%%%%%%%%%%%%%%%%%%%%%%%%%%%%%%%%%%%%%%%%%%%%%%%%%
\subsection*{Property 1: Sparsity of the gradient vector}
%%%%%%%%%%%%%%%%%%%%%%%%%%%%%%%%%%%%%%%%%%%%%%%%%%%%%%%%%%%%%%%%%%%%%%%%%%%%%%%%%%%%%%%%%
%\textit{Property 1: Sparsity of the gradient vector} 
For a fixed number of variables $n$, the gradient vector has a maximum of $4n$ non-zero entries for any number of basis functions $K \geq 4$: 

$$ 
\norm{ \grad_{\vec{\mathbf{\theta}}} f(\vec{\mathbf{x}} )}_{0} 
= \sum_{i=1}^{Kn}  d_{Hamming} \left(\frac{\partial f}{\partial \theta_{i}} (\vec{\mathbf{x}}),0  \right)
\leq 4 n
\; \forall \; \vec{\mathbf{x}} \in D(f) \subset R^{n}
$$

The sparsity of the multi-variable SAM model follows from the sparsity of each single-variable function used. SAM is robust to catastrophic forgetting.

%%%%%%%%%%%%%%%%%%%%%%%%%%%%%%%%%%%%%%%%%%%%%%%%%%%%%%%%%%%%%%%%%%%%%%%%%%%%%%%%%%%%%%%%%
\subsection*{Property 2: Gradient flow attenuation}
%%%%%%%%%%%%%%%%%%%%%%%%%%%%%%%%%%%%%%%%%%%%%%%%%%%%%%%%%%%%%%%%%%%%%%%%%%%%%%%%%%%%%%%%%

For a fixed number of variables $n$, the gradient vector has bounded L1 norm for any number of basis functions $K \geq 4$:

$$ 
\norm{ \grad_{\vec{\mathbf{\theta}}} f( \vec{\mathbf{x}} )}_{1} 
= \sum_{i=1}^{Kn}  \left| \frac{\partial f}{\partial \theta_{i}} ( \vec{\mathbf{x}} ) \right| 
< 4Un
\; \forall \; \vec{\mathbf{x}} \in D(f)
$$

The bounded norm for SAM follows from the single-variable case. SAM is numerically stable during training and well-behaved in the limit of arbitrarily many basis functions.

% 
%%%%%%%%%%%%%%%%%%%%%%%%%%%%%%%%%%%%%%%%%%%%%%%%%%%%%%%%%%%%%%%%%%%%%%%%%%%%%%%%%%%%%%%%%
\subsection*{Property 3: Distal orthogonality}
%%%%%%%%%%%%%%%%%%%%%%%%%%%%%%%%%%%%%%%%%%%%%%%%%%%%%%%%%%%%%%%%%%%%%%%%%%%%%%%%%%%%%%%%%

%\textit{Property 3: Distal orthogonality}

For any spline additive model, there exists a $\delta>0$ such that:
$$
 \min_{j=1, \dots , n }
\{ |x_{j} - y_{j}| \} > \delta 
\implies  
\langle
\grad_{\vec{\mathbf{\theta}}} f(\vec{\mathbf{x}}) 
, 
\grad_{\vec{\mathbf{\theta}}} f(\vec{\mathbf{y}})
\rangle
= 0 \; 
\forall \; \vec{\mathbf{x}},\vec{\mathbf{y}} \in D(f) \subset R^{n}
$$

The distal orthogonality follows from the single-variable case. Two points that sufficiently differ in each input-variable have orthogonal parameter gradients. It is worth mentioning that the condition resembles a cross-like region in two variables, and planes that intersect in higher dimensions. Distal orthogonality means SAM is reasonably robust to catastrophic forgetting.

%%%%%%%%%%%%%%%%%%%%%%%%%%%%%%%%%%%%%%%%%%%%%%%%%%%%%%%%%%%%%%%%%%%%%%%%%%%%%%%%%%%%%%%%%
\subsection*{General Overview of SAM}
%%%%%%%%%%%%%%%%%%%%%%%%%%%%%%%%%%%%%%%%%%%%%%%%%%%%%%%%%%%%%%%%%%%%%%%%%%%%%%%%%%%%%%%%%

SAMs also exhibit stratification and inherent memory retention that is robust, but not perfect since the overlapping regions are not as localised as in the single-variable case. SAMs can be implemented as neural networks, as illustrated in Figure~\ref{fig:SAM_implementation}. SAMs are not a universal function approximation scheme, there are continuous multi-variable functions that cannot be expressed as a sum of single variable functions. However, for problems where the manifold hypothesis is true and data lies on a very low-dimensional manifold, SAMs can be sufficient. Additionally, SAMs could be well-suited for use with kernel techniques, or the Fourier transform of the input. Recurrent neural networks and reservoir computers may also benefit from the robust nature of SAMs.

%%%%%%%%%%%%%%%%%%%%%%%%%%%%%%%%%%%%%%%%%%%%%%%%%%%%%%%%%%%%%%%%%%%%%%%%%%%%%%%%%%%%%%%%%
%\section{Kolmogorov-Arnold Representation Theorem}\label{sec:kolmogorov}
\section{KASAM: Universal Function Approximation}\label{sec:KASAM}
%%%%%%%%%%%%%%%%%%%%%%%%%%%%%%%%%%%%%%%%%%%%%%%%%%%%%%%%%%%%%%%%%%%%%%%%%%%%%%%%%%%%%%%%%

The Kolmogorov-Arnold representation shows how any multi-variable function can be represented with single-variable functions~\cite{kolmogorov1957representation,kolmogorovRevisited}. The theorem states that any continuous multi-variable function $y$ on the unit hyper-cube with $n$ input-variables $x_{p}$ can be exactly represented with continuous single-variable functions $\Phi_{q}$, $\phi_{q,p}$ of the form: 

$$
y(\vec{\mathbf{x}}) 
= y(x_{1},...,x_{n}) 
= \sum^{2n}_{q=0} \Phi_{q} \left( \sum^{n}_{p=1} \phi_{q,p}( x_{p} ) \right)
$$

The representation theorem is not an approximation - it is exact~\cite{kolmogorov1957representation,kolmogorovRevisited}. Furthermore, there is no mention of the computability or learnability of the representation in the theorem. It is also notable that the summation in the representation is finite, and not arbitrarily large or infinite as with Taylor series. It is also unlike the Universal Function Approximation theorems for arbitrarily wide and arbitrarily deep neural networks, which also correspond to summing together arbitrarily many terms. The core building block for the Kolmogorov-Arnold representation theorem is continuous single-variable functions.

% 
%%%%%%%%%%%%%%%%%%%%%%%%%%%%%%%%%%%%%%%%%%%%%%%%%%%%%%%%%%%%%%%%%%%%%%%%%%%%%%%%%%%%%%%%%
%\section{KASAM: Universal Function Approximation}\label{sec:traumtensor}
%%%%%%%%%%%%%%%%%%%%%%%%%%%%%%%%%%%%%%%%%%%%%%%%%%%%%%%%%%%%%%%%%%%%%%%%%%%%%%%%%%%%%%%%%

The approximation scheme for KASAM stems from the arbitrary density of basis functions to approximate each single-variable function. A sum of single-variable functions $g_{p}$ is added to the expression. Inspiration was taken from ResNet arcitectures to make optimisation easier~\citep{resnetpaper}. The Kolmogorov-Arnold representation theorem can be extended with a ResNet-style residual skip-connection for some constant $\lambda$ to obtain the definition for KASAM. The function approximator $f$ replaces the summation over $2n$ with a summation over some $N$. The functions $ h_{q,p}$ and $g_{j}$ are uniform B-spline functions. The outer functions are given by $H_{q}(z) = b_{q} ( \sigma(z))$ where $\sigma$ is the sigmoid function that maps any real number to the unit interval. The functions $s_{q}$ are uniform B-spline functions. The overall expression for Kolmogorov-Arnold Spline Additive Model (KASAM) is given by:

$$
f(\vec{\mathbf{x}})
= \sum^{N}_{q=1} H_{q} \left( \sum^{n}_{p=1} h_{q,p}( x_{p} ) \right)
    + \lambda \sum^{N}_{q=1}
    \left( \sum^{n}_{p=1} h_{q,p}( x_{p} ) \right) 
    + \sum^{n}_{j=1} g_{j}( x_{j} )
$$

There is a one-to-one correspondence between the exact representation and the structure of KASAM (assuming $N = 2n$). Keep in mind that for $\lambda \neq 0$ one can choose the functions $g_{p}$ such that it cancels out with the residual term, yielding the original representation theorem. 

If an arbitrary density of basis functions are used, then one can approximate the exact target function. It is hoped that using cubic B-splines and SAM would give rise to a model that is easy to implement and train. Unfortunately the analytical guarantees that SAM possesses do not hold for KASAM in general.

A more compact vector notation can be given, replacing $\lambda$ with a constant matrix or linear projection $T$:

$$
f(\vec{\mathbf{x}})
= H \left(  \vec{\mathbf{h}}(\vec{\mathbf{x}}) \right)
    + T \vec{\mathbf{h}}(\vec{\mathbf{x}})
    + g(\vec{\mathbf{x}})
$$

\begin{figure}[!h]
\centering
\noindent
\includegraphics[width=\textwidth]{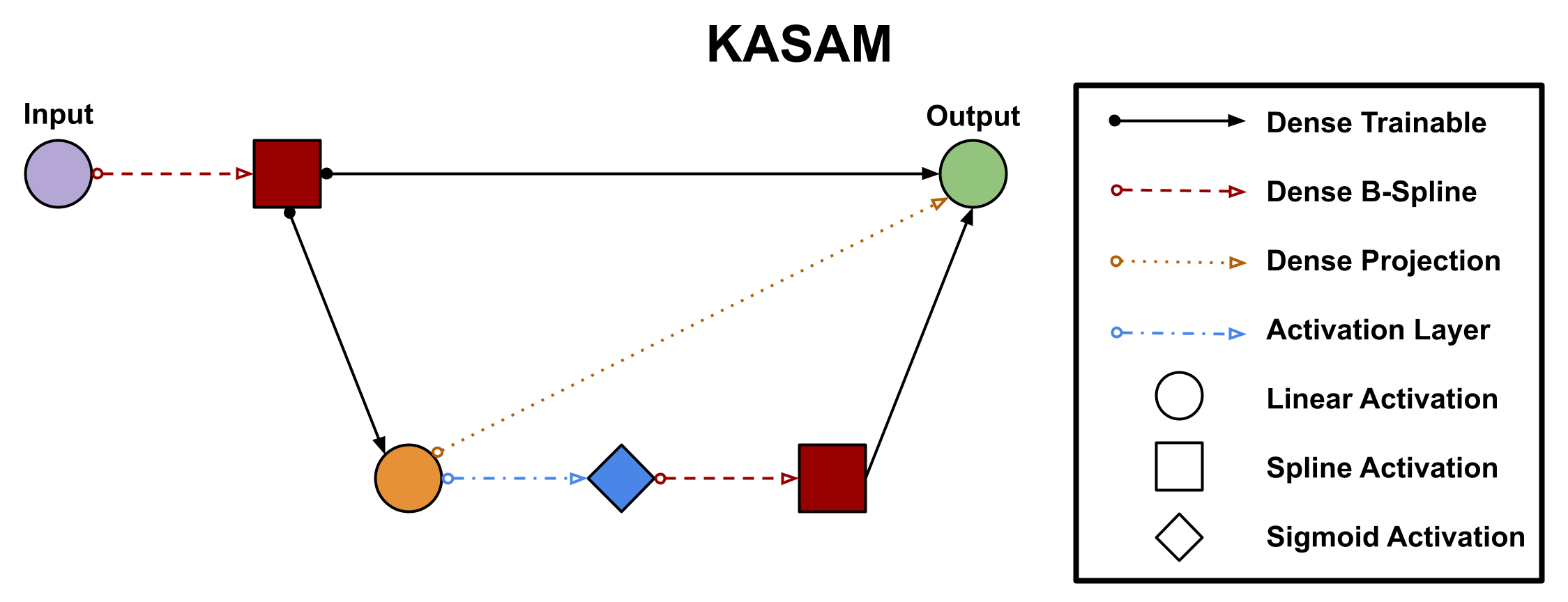}
\caption{\small Structure of KASAM. The computational graph represents layer(s) as nodes. The SAM module is used extensively in KASAM. One branch is passed through a sigmoid function, and is then passed to another SAM layer (implementing the exterior functions). There is a residual skip-connection implemented with a constant linear layer between the interior SAM module and the output of the model.}
\label{fig:fig_graph_KASAM_stucture}
\end{figure}

KASAM uses SAM modules throughout to approximate all sums of single-variable functions, as shown in Figure~\ref{fig:fig_graph_KASAM_stucture}. The TensorFlow implementation and prototype is a proof of concept and can be seen in the linked \href{https://github.com/hpdeventer/KASAM}{GitHub repository}\footnote{\href {https://github.com/hpdeventer/KASAM}{https://github.com/hpdeventer/KASAM}}. The focus was on developing a working prototype, and not computational efficiency. Future research into more efficient implementations would be ideal.

%%%%%%%%%%%%%%%%%%%%%%%%%%%%%%%%%%%%%%%%%%%%%%%%%%%%%%%%%%%%%%%%%%%%%%%%%%%%%%%%%%%%%%%%%
\subsection*{General Overview of KASAM}
%%%%%%%%%%%%%%%%%%%%%%%%%%%%%%%%%%%%%%%%%%%%%%%%%%%%%%%%%%%%%%%%%%%%%%%%%%%%%%%%%%%%%%%%%

%KASAM is a universal function approximator, unlike SAM which has limitied expressive power.
KASAM is a universal function approximator. KASAM can be implemented as a special type of artificial neural network with specifically chosen weights and activation functions, and trainable weights that can be optimised with gradient descent algorithms. The KASAM neural network that is constructed from SAM might inherit some memory retention, although it is not as predictably and reliably robust to catastrophic forgetting as SAMs. Stratification could hinder KASAM's generalisation on certain tasks, and it is not obvious how to manage this potential weakness. To reduce catastrophic forgetting, a pseudo-rehearsal training technique can be used. Further implementation details can be found in Section~\ref{sec:method} and the Github repository.
 
%%%%%%%%%%%%%%%%%%%%%%%%%%%%%%%%%%%%%%%%%%%%%%%%%%%%%%%%%%%%%%%%%%%%%%%%%%%%%%%%%%%%%%%%%
\section{Pseudo-Rehearsal Training Techniques}\label{sec:pseudo}
%%%%%%%%%%%%%%%%%%%%%%%%%%%%%%%%%%%%%%%%%%%%%%%%%%%%%%%%%%%%%%%%%%%%%%%%%%%%%%%%%%%%%%%%%

Numerous review, rehearsal and pseudo-rehearsal techniques exist~\citep{robins1995catastrophic}. Some use generative models like GANs~\citep{shin2017continual}. The expected risk has the familiar form:

$$ R(f) = \int \ell (f(x),y) \mathrm{d}P(x,y)$$

Pseudo-rehearsal explicitly refers back to the model's previous state, and is similar to the ideas related to generative replay~\citep{shin2017continual}. The distribution $\mathcal{P}(z)=\mathcal{P}_{D(f)}$ of the input data with $z \sim \mathcal{P}_{D(f)}$ could be given by a generative model or a uniform distribution $z \sim U_{D(f)}$ over the domain of $f$ denoted $D(f)$. Using a uniform distribution is less computationally demanding than training a generative model. The listed integral elicits memory. The mixing coefficient $\rho \in \left[0,1 \right]$ controls novel learning and memory retention. This technique is given by the discrete-time functional of the form:

$$ 
R(f_{t+1}) = 
\rho \int \ell (f_{t+1}(x),y) \mathrm{d}P(x,y) + (1-\rho)\int \ell (f_{t+1}(z),f_{t}(z))\mathrm{d}\mathcal{P}(z)
$$

\begin{figure}[!h]
\centering
\noindent
\includegraphics[width=0.6\textwidth]{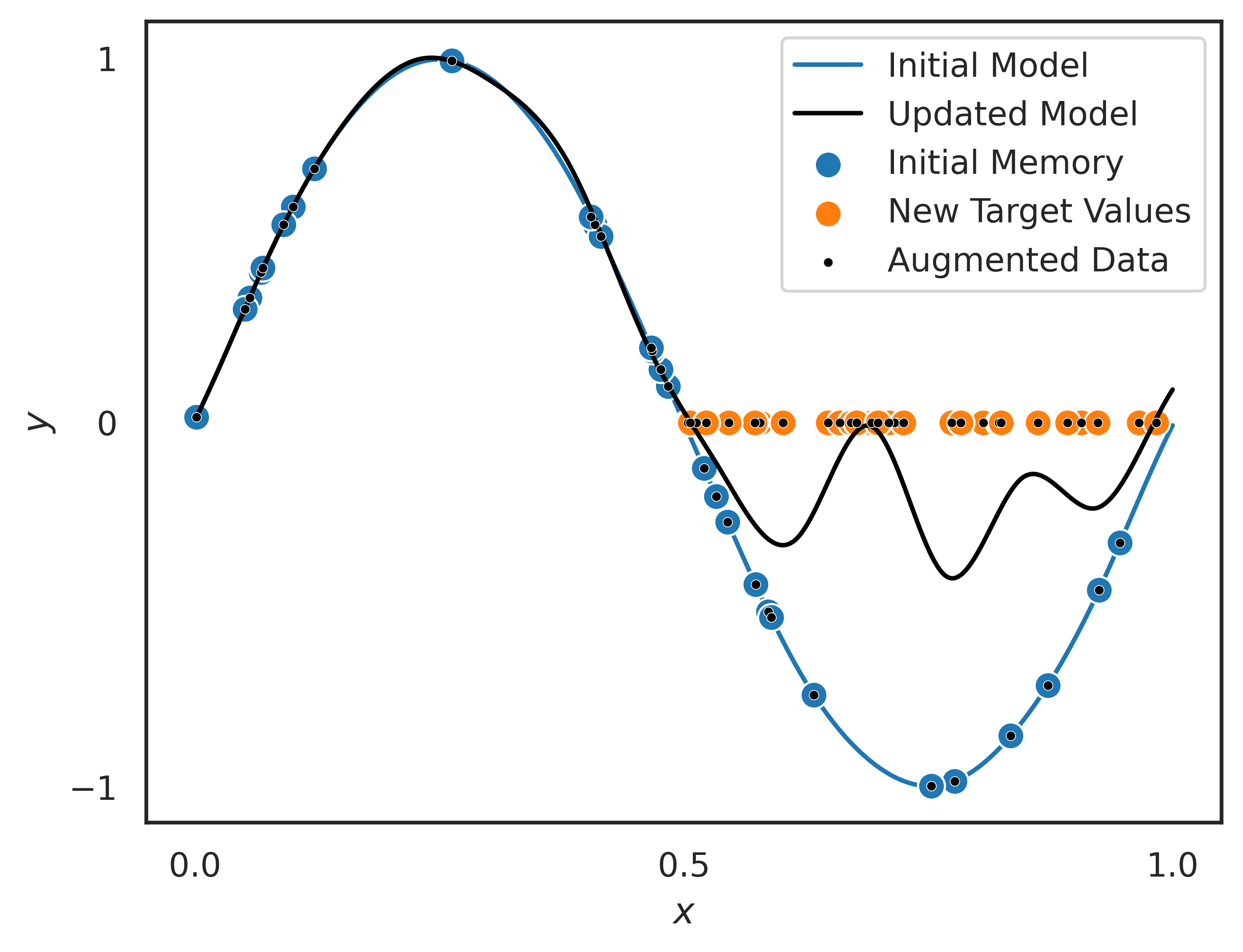}
\caption{\small Pseudo-rehearsal is demonstrated. The training data is augmented with data sampled from the model's initial input-output values. Pseudo-rehearsal exploits the model's memory of the previous tasks to retain memory and change only in regions where the new data is present.}
\label{fig:fig_label_reveries}
\end{figure}

The key idea is to minimise the loss on new data, subject to the constraint that the model retains its input-output mapping over the rest of its domain. The same functional can be seen in the paper~\citep{shin2017continual}. This could mitigate catastrophic forgetting and is an iterative process that may or may not converge depending on the choice of loss function $\ell$, model $f$, and the target values. A simple version of pseudo-rehearsal is demonstrated in Figure~\ref{fig:fig_label_reveries}.

%%%%%%%%%%%%%%%%%%%%%%%%%%%%%%%%%%%%%%%%%%%%%%%%%%%%%%%%%%%%%%%%%%%%%%%%%%%%%%%%%%%%%%%%%
\section{Methodology}\label{sec:method}
%%%%%%%%%%%%%%%%%%%%%%%%%%%%%%%%%%%%%%%%%%%%%%%%%%%%%%%%%%%%%%%%%%%%%%%%%%%%%%%%%%%%%%%%%

%\subsection*{Description of Models}

Four models were considered for experimental evaluation. The first was a SAM based model with no additional regularisation, called \textit{SAM}. The second model was a feed-forward ANN with the same structure and activation functions as KASAM, but all the model parameters were trainable and randomly initialised instead of set to predefined constants, further referred to as simply \textit{ANN}. The third was KASAM, with some parameters being trainable and others being fixed to correctly implement cubic B-splines, named \textit{KASAM}. (Note: the ANN is capable of learning an identical clone of KASAM, but such eventuality is unlikely given the random initialisation of the ANN). The fourth model was KASAM in combination with the pseudo-rehearsal (PR) data-augmentation technique, referred to as \textit{KASAM+PR}.

\subsubsection*{SAM}
The SAM model is a scalar-valued function that maps a two-dimensional input to a one-dimensional output. A density of 32 basis functions was chosen for each input variable. The SAM model chosen implements:

$$
f(\vec{\mathbf{x}})  
= \sum^{2}_{j=1} f_{j}( x_{j} ) 
= \sum^{2}_{j=1} \sum_{i=1}^{32}   \theta_{i,j} S_{i,j}(  x_{j} )
= \sum^{2}_{j=1} \sum_{i=1}^{32}   \theta_{i,j} S(w_{i,j}x_{j} + b_{i,j})
$$

\begin{figure}[!h]
\centering
\noindent
\includegraphics[width=\textwidth]{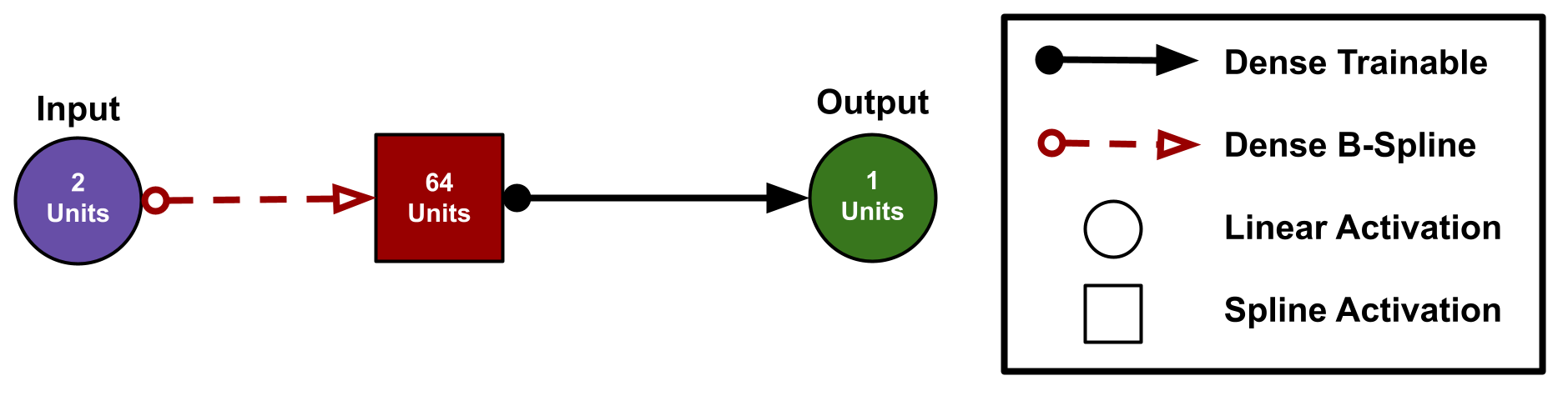}
\caption{\small Structure of SAM.}
\label{fig:fig_SAM_model_specifics}
\end{figure}

SAM has a reasonably simple structure shown in Figure~\ref{fig:fig_SAM_model_specifics}. The inputs were two-dimensional and the output was one-dimensional. The density of basis functions for each input-variable is 32, so there are 64 neural units in the hidden layer. The activation function and weights were chosen to correctly implement B-spline basis functions. The final linear layer is trainable - each basis function is multiplied by its trainable parameter and summed together to give output of SAM. 

\subsubsection*{KASAM}
The KASAM model is a scalar valued function that maps a two dimensional input to a one-dimensional output. It was decided to use three exterior functions such that:

$$
f(\vec{\mathbf{x}})
= \sum^{3}_{q=1} H_{q} \left( \sum^{2}_{p=1} h_{q,p}( x_{p} ) \right)
    + \lambda \sum^{3}_{q=1}
    \left( \sum^{2}_{p=1} h_{q,p}( x_{p} ) \right) 
    + \sum^{2}_{j=1} g_{j}( x_{j} )
$$

The structure of KASAM is a generalisation of SAM. One branch is the same as SAM. For technical and practical reasons to make optimisation easier it was necessary to use a mixture of different densities of basis functions. There are densities of 4, 8, 16 and 32 basis functions for each input variable. The maximum density of 32 corresponds to the same expressive power of the mentioned SAM model also with a density of 32 basis functions. Adding together all the densities of basis functions in KASAM gives a total 60 basis functions for each variable. The two input variables have 120 basis functions or neural units in total. The three hidden variables have 180 basis functions or neural units in total - this implements three exterior functions each with their own input.

Most of the connections represent fully connected dense layers. The only exception is the activation layer that only applies an element-wise sigmoid to each input value, with no trainable parameters. The layers with trainable parameters are indicated with solid black arrows. The constant parameters or weights are not solid black arrows.

\begin{figure}[!h]
\centering
\noindent
\includegraphics[width=\textwidth]{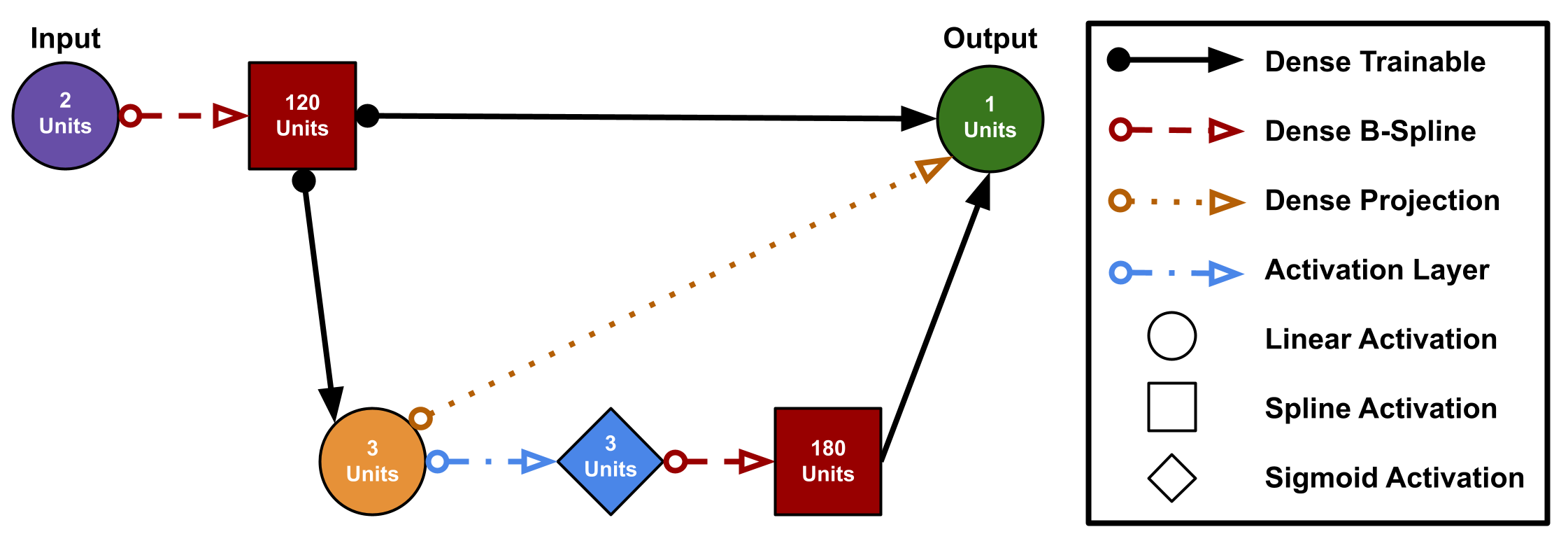}
\caption{\small Structure of KASAM.}
\label{fig:fig_SAM_Two_var}
\end{figure}

KASAM is fully capable of representing the SAM model exactly, if the appropriate parameters were chosen or zeroed.

\subsubsection*{KASAM+PR}
KASAM+PR has the same structure as KASAM. The only difference is that a pseudo-rehearsal data augmentation is utilised in training. The training data for the second task is mixed with input-output values of the model after it was trained on Task 1 (but before being trained on Task 2). The rehearsal dataset is constructed from 10000 uniformly sampled points of two dimensions on the unit interval. The target values for rehearsal are predicted by the model itself, using the stored memory of the previous data in the model. The augmented dataset is 10000 points randomly sampled with $50\%$ probability of choosing either a rehearsal data point or training data point for Task 2.

\subsubsection*{ANN}

\begin{figure}[!h]
\centering
\noindent
\includegraphics[width=\textwidth]{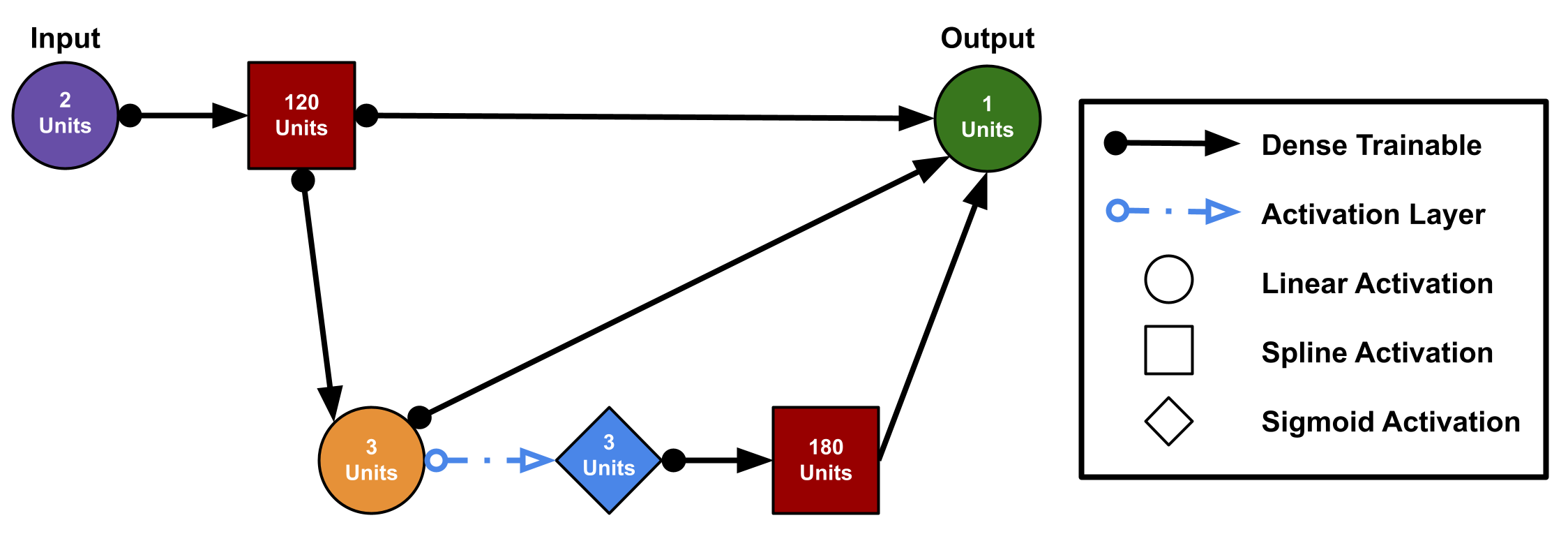}
\caption{\small Structure of the ANN model.}
\label{fig:fig_ANN_model_specifics}
\end{figure}

The ANN model has the same structure as KASAM. The ANN model is similar to more commonly used initialisation methods. The ANN model has randomly initialised and trainable parameters, whereas KASAM has many specifically chosen parameters that are not trainable. It is possible for ANN to implement KASAM, but it is unlikely. ANN does not neatly decompose into single-variable function because some of the parameters in the dense layers are not zero. All that remains is the structure as seen in Figure~\ref{fig:fig_ANN_model_specifics}, but all parameters are trainable and randomly initialised with the default values for TensorFlow.

The ANN model was chosen to show how randomly initialised parameters compare to the specifically chosen parameters for KASAM. Given appropriate random initialisation, hyper-parameters, optimizer, and excessive data, it is possible to get the same performance from ANN as with KASAM. Most ANNs are very sensitive to many externally chosen values, and require a lot of fine-tuning for acceptable performance.

%%%%%%%%%%%%%%%%%%%%%%%%%%%%%%%%%%%%%%%%%%%%%%%%%%%%%%%%%%%%%%%%%%%%%%%%%%%%%%%%%%%%%%%%%
 
\subsection{Experiments}
%%%%%%%%%%%%%%%%%%%%%%%%%%%%%%%%%%%%%%%%%%%%%%%%%%%%%%%%%%%%%%%%%%%%%%%%%%%%%%%%%%%%%%%%%

The target functions for Experiments A, B and C were chosen to demonstrate the expressive power and limitation of all four models. The target function for Experiment A is a sum of single-variable functions which is easily expressed (in theory) by the SAM, KASAM and KASAM+PR models. The target function for Experiment B is a modified version of a difficult to learn two variable function that is the sum of single variable and multi-variable functions~\citep{malan_cleghorn}. The target function for Experiment C is a product of periodic functions which is impossible for SAM to represent.

All models and experiments were performed with Python and TensorFlow. The loss function chosen for training and evaluation in all presented experiments is mean absolute error (MAE). The training data set and test set in all experiments had $10 000$ data points. Gaussian noise of variance $0.05$ was added to all training and test data target values. The test set was also used as a validation set to quantify the test error during training. All models were trained with a learning rate of $0.001$ with the Adam optimizer. All models and experiments used batch sizes of 100 during training. 

%%%%%%%%%%%%%%%%%%%%%%%%%%%%%%%%%%%%%%%%%%%%%%%%%%%%%%%%%%%%%%%%%%%%%%%%%%%%%%%%%%%%%%%%%
\subsubsection*{Task 1}
%%%%%%%%%%%%%%%%%%%%%%%%%%%%%%%%%%%%%%%%%%%%%%%%%%%%%%%%%%%%%%%%%%%%%%%%%%%%%%%%%%%%%%%%%
The training and test sets were sampled uniformly from the Task 1 target functions over the domain $\left[0.,1. \right]^{2}$, with Gaussian noise added to the target values. All models were trained for $200$ epochs. The Task 1 target functions for experiments A, B, and C are given by:

\begin{equation*} \label{eq1}
\begin{split}
Y_{A}(x_{1},x_{2}) &= \cos{(4\pi x_{1})}\exp(-(2x_{1}-1)^{2})+\sin{(\pi x_{2})}                     \\
Y_{B}(x_{1},x_{2}) &= 2\exp(-\sum_{i=1}^{2} (10 x_{i}-5)^{2}) + \sum_{i=1}^{2}\sin^{2}(10x_{i}-5)  \\
Y_{C}(x_{1},x_{2}) &= 1 + \cos(20x_{1}-10) \cos(20x_{2}-10)                                         
\end{split}
\end{equation*}

%%%%%%%%%%%%%%%%%%%%%%%%%%%%%%%%%%%%%%%%%%%%%%%%%%%%%%%%%%%%%%%%%%%%%%%%%%%%%%%%%%%%%%%%%
\subsubsection*{Task 2}
%%%%%%%%%%%%%%%%%%%%%%%%%%%%%%%%%%%%%%%%%%%%%%%%%%%%%%%%%%%%%%%%%%%%%%%%%%%%%%%%%%%%%%%%%

The test sets were sampled uniformly from the Task 2 target functions over the domain $\left[0.,1. \right]^{2}$, with Gaussian noise added to the target values. The training sets were sampled uniformly over the domain $\left[0.45,0.55 \right]^{2}$, and target values of zero with added Gaussian noise. All models were trained for $20$ epochs. The Task 2 target functions for experiments A, B, and C are given by:

$$ Y'_{A}(x_{1},x_{2}) =\begin{cases} 
      0 &  0.45 < x_{1} < 0.55, \text{ and } 0.45 < x_{2} < 0.55          \\
      Y_{A}(x_{1},x_{2}) & \text{otherwise.} 
   \end{cases}
$$

$$ Y'_{B}(x_{1},x_{2}) =\begin{cases} 
      0 &  0.45 < x_{1} < 0.55, \text{ and } 0.45 < x_{2} < 0.55          \\
      Y_{B}(x_{1},x_{2}) & \text{otherwise.} 
   \end{cases}
$$

$$ Y'_{C}(x_{1},x_{2}) =\begin{cases} 
      0 &  0.45 < x_{1} < 0.55, \text{ and } 0.45 < x_{2} < 0.55          \\
      Y_{C}(x_{1},x_{2}) & \text{otherwise.} 
   \end{cases}
$$

%%%%%%%%%%%%%%%%%%%%%%%%%%%%%%%%%%%%%%%%%%%%%%%%%%%%%%%%%%%%%%%%%%%%%%%%%%%%%%%%%%%%%%%%%
\section{Empirical Results}\label{sec:exp}
%%%%%%%%%%%%%%%%%%%%%%%%%%%%%%%%%%%%%%%%%%%%%%%%%%%%%%%%%%%%%%%%%%%%%%%%%%%%%%%%%%%%%%%%%

\subsection{Experiment A}

The mean and standard deviation of the test MAE over thirty independent trials for each model are shown in Table~\ref{table:A_results_averaged}. The null hypothesis for each pair-wise comparison between models is that they have indistinguishable test errors (threshold is $p<0.0001$). The p-values were calculated from raw data. 

\begin{figure} [!h]
     \centering
     \begin{subfigure}[!h]{0.49\textwidth}
         \centering
         \includegraphics[width=\textwidth]{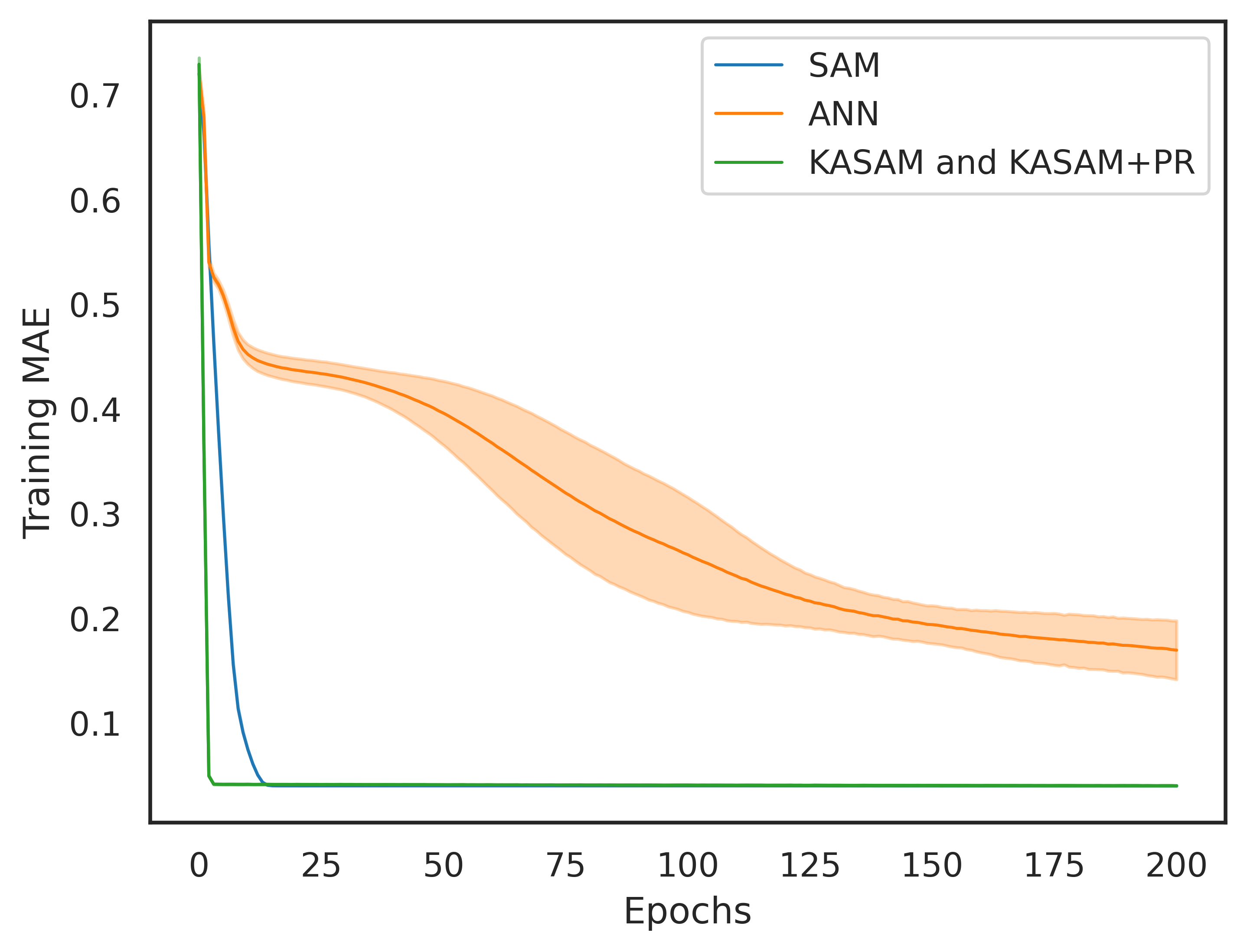}
         \caption{Training loss curve.}
         \label{fig:A_task_1_training_plot}
     \end{subfigure}
     \hfill
     \begin{subfigure}[!h]{0.49\textwidth}
         \centering
         \includegraphics[width=\textwidth]{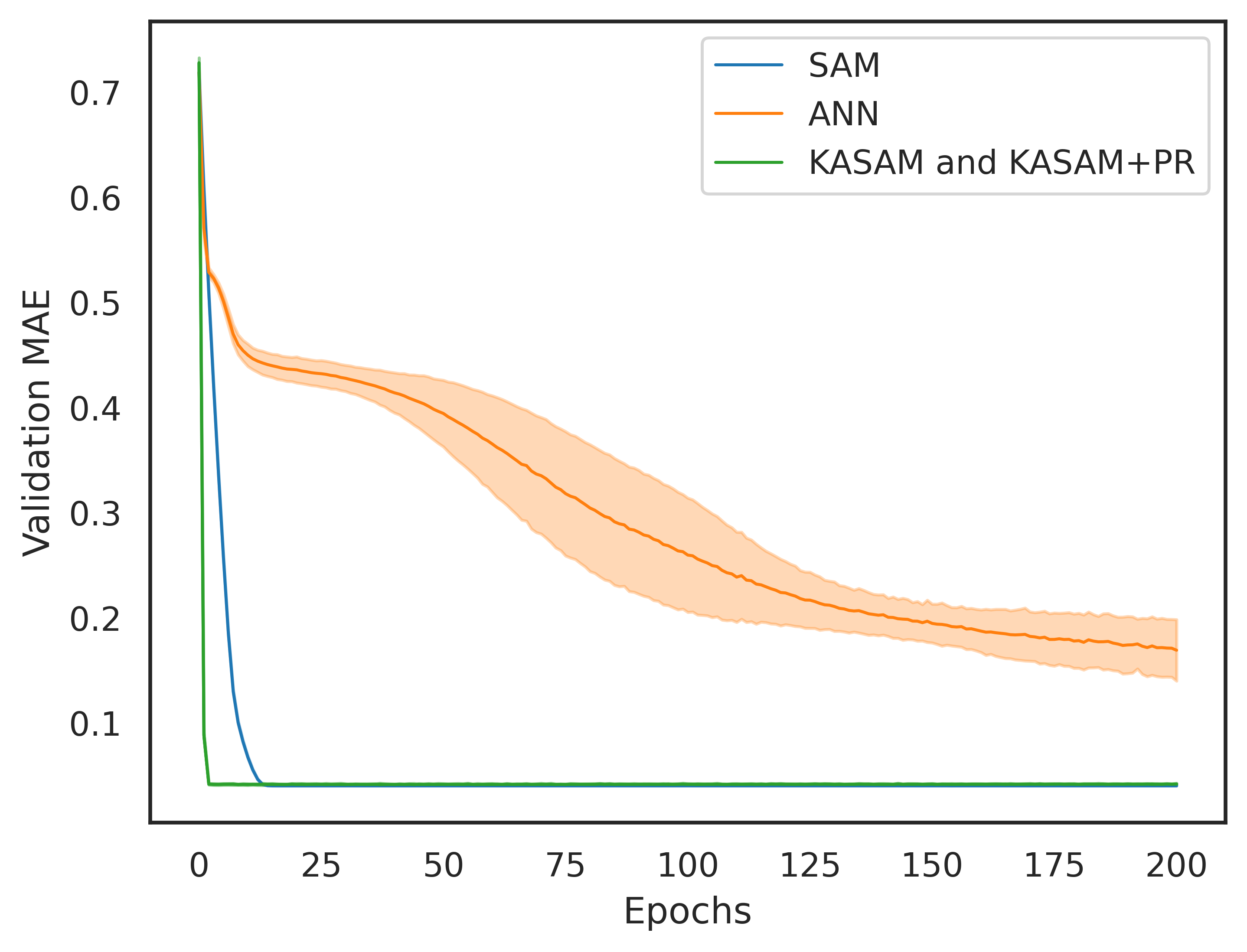}
         \caption{Validation loss curve.}
         \label{fig:A_task_1_validation_plot}
     \end{subfigure}
        \caption{\small 
Experiment A: Task 1 training and validation loss during training. All models were trained on $10000$ training data-points. The initial loss before training is shown at the zeroth epoch.}
        \label{fig:A_task_1_training_validation_plot}
\end{figure}

Task 1 indicated that KASAM and KASAM+PR have indistinguishable test errors, and the null hypothesis was accepted ($p=0.1822$). All other pair-wise comparisons for Task 1 indicated distinguishable test errors, and the null hypothesis was rejected ($p<0.0001$). Rounding the test MSE in Task 1 to a few decimal places shows that SAM, KASAM, and KASAM+PR have practically the same test error as shown in Table~\ref{table:A_results_averaged}. The training and validation MAE during training is shown in Figure~\ref{fig:A_task_1_training_validation_plot}. The test sets were used for validation as well. The SAM, KASAM and KASAM+PR models easily learned the target function, and reasonably quickly. The ANN model struggled to learn the Task 1 target function for experiment A as seen in Figure~\ref{fig:A_task_1_training_validation_plot}. 

\begin{figure} [!h]
     \centering
     \begin{subfigure}[!h]{0.49\textwidth}
         \centering
         \includegraphics[width=\textwidth]{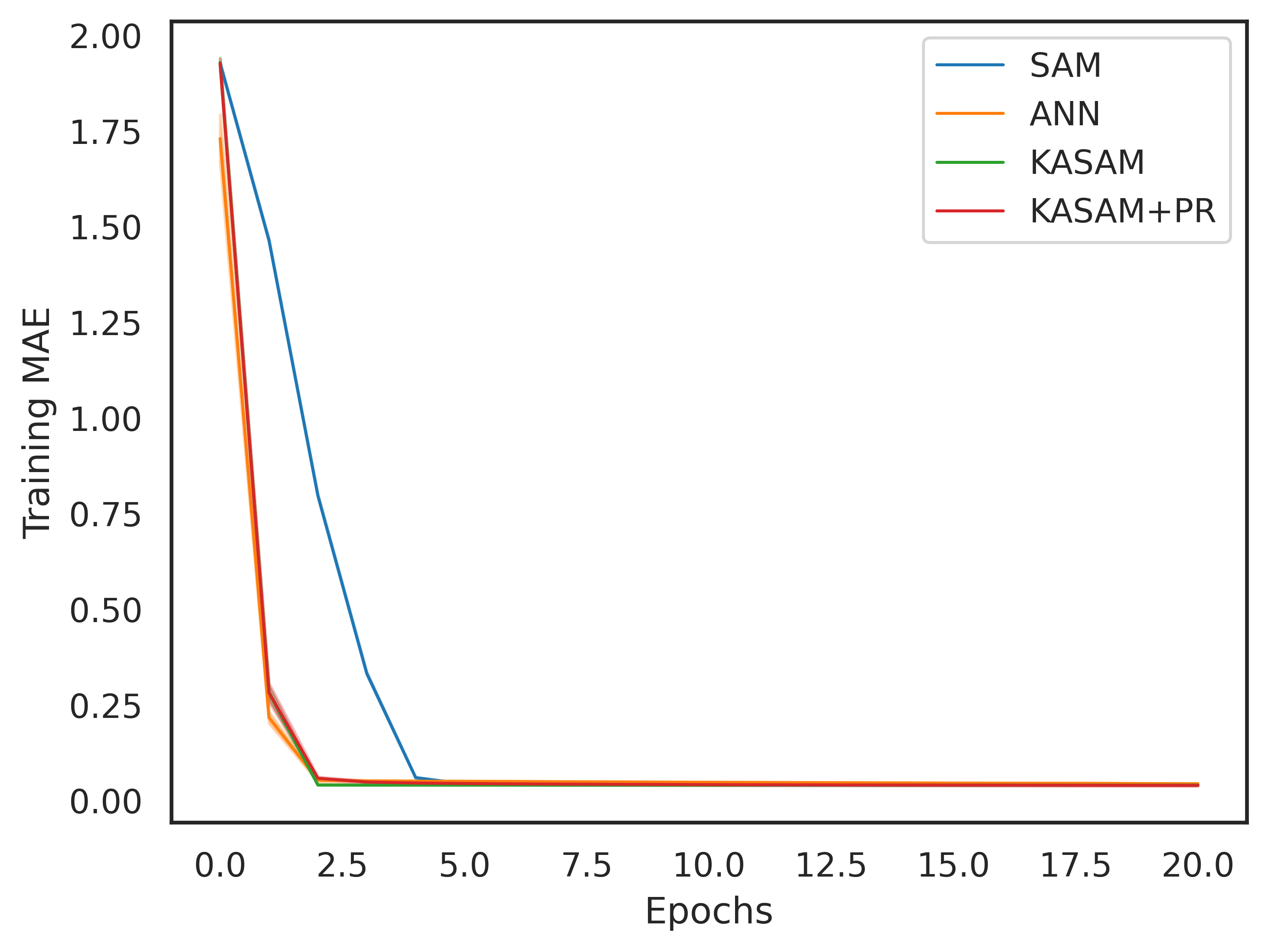}
         \caption{Training loss curve.}
         \label{fig:A_task_2_training_plot}
     \end{subfigure}
     \hfill
     \begin{subfigure}[!h]{0.49\textwidth}
         \centering
         \includegraphics[width=\textwidth]{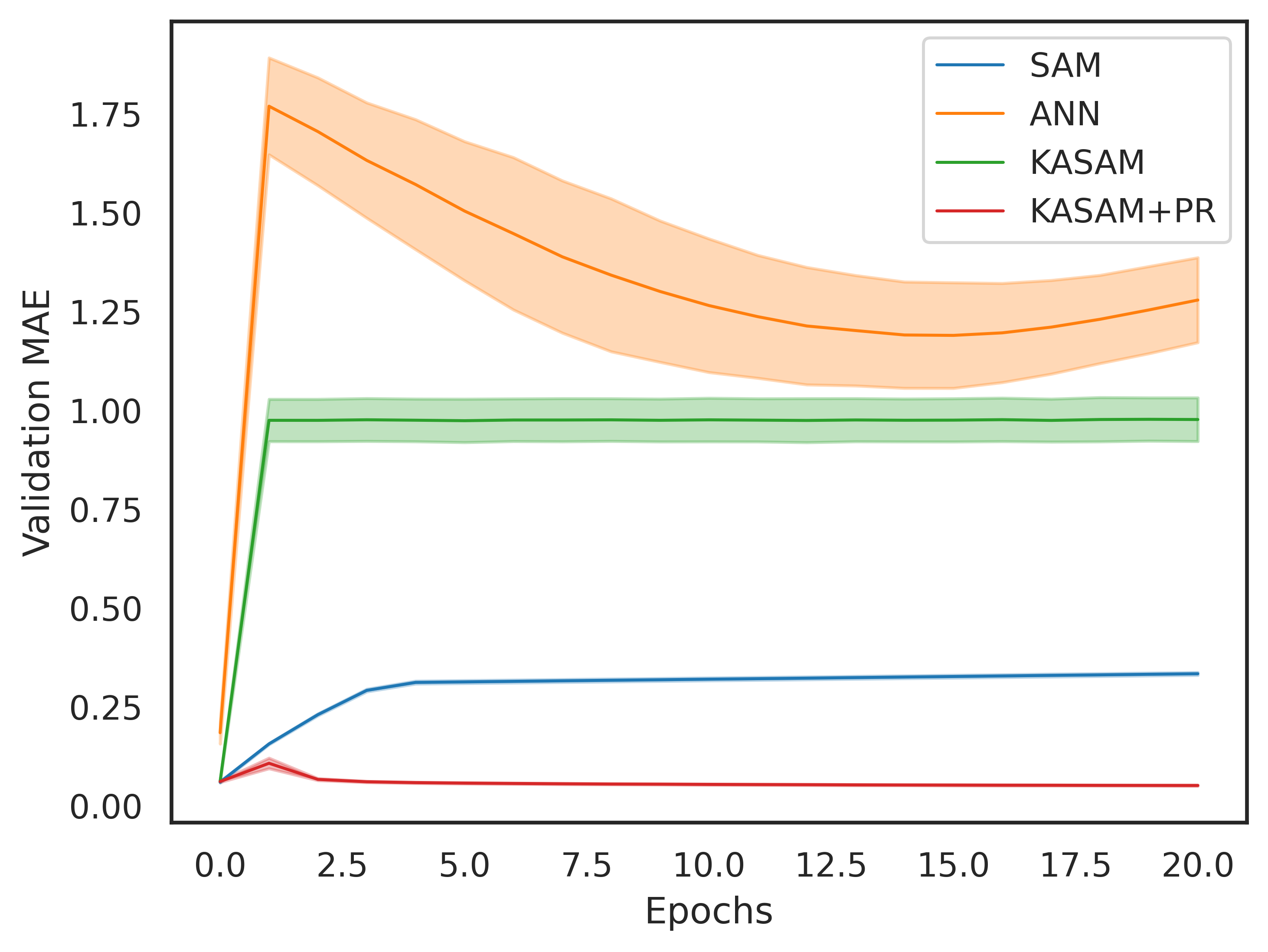}
         \caption{Validation loss curve.}
         \label{fig:A_task_2_validation_plot}
     \end{subfigure}
        \caption{\small 
Experiment A: Task 2 training and validation loss during training. All models were trained on $10000$ training data-points. The initial loss before training is shown at the zeroth epoch.}
        \label{fig:A_task_2_training_validation_plot}
\end{figure}

Task 2 indicated that all four models had distinguishable test errors, and the null hypothesis was rejected ($p<0.0001$) for each pair-wise comparison. KASAM+PR had the best test MAE indicating the benefit of using pseudo-rehearsal techniques with KASAM as shown in Table~\ref{table:A_results_averaged}. The SAM model had the second best performance on Task 2 with some memory retention. The KASAM model alone had the third best performance on Task 2, indicating marginal memory retention. The ANN model had the worst performance and suffered catastrophic forgetting that severely impedes its performance compared to the other models in Table~\ref{table:A_results_averaged}. The training and validation MAE during training is shown in Figure~\ref{fig:A_task_2_training_validation_plot}. The test sets were used for validation as well. All models had similar training loss curves in Figure~\ref{fig:A_task_2_training_validation_plot}. The validation loss curves displayed interesting dynamics during training Figure~\ref{fig:A_task_2_training_validation_plot}. KASAM+PR performed the best of all models, and pseudo-rehearsal limited catastrophic forgetting and allowed the model to improve after initially degrading in performance. The SAM model degraded in performance and plateaued with little variance. The KASAM and ANN models had the worst performance with a lot of variance in validation MAE. The ANN validation loss curves exhibited oscillatory behaviour possibly due to the use of Adam as an optimizer, as shown in Figure~\ref{fig:A_task_2_training_validation_plot}. 

\begin{table}[!h]
\centering
\begin{tabular}{|c c c|} 
 \hline
                    & Task 1 MAE                & Task 2 MAE  \\ [0.5ex] 
 \hline\hline
 SAM                & \bf{0.040 (0.000)}    & 0.334 (0.004)          \\ 
 ANN                & 0.169 (0.029)         & 1.279 (0.107)          \\ 
 KASAM              & 0.042 (0.001)         & 0.977 (0.055)          \\ 
 KASAM+PR           & 0.042 (0.001)         & \bf{0.051 (0.001)}     \\ [0.5ex] 
 \hline
\end{tabular}
\caption{Experiment A: final test mean absolute error (MAE) for Task 1 and Task 2 averaged over 30 trials, rounded to two decimal places.}
\label{table:A_results_averaged}
\end{table}

%%%%%%%%%%%%%%%%%%%%%%%%%%%%%%%%%%%%%%%%%%%%%%%%%%%%%%%%%%%%%%%%%%%%%%%%%%%%%%%%%%%%%%%%%

\begin{figure} [!h]
     \centering
     \begin{subfigure}[!h]{0.9\textwidth}
         \centering
         \includegraphics[width=\textwidth]{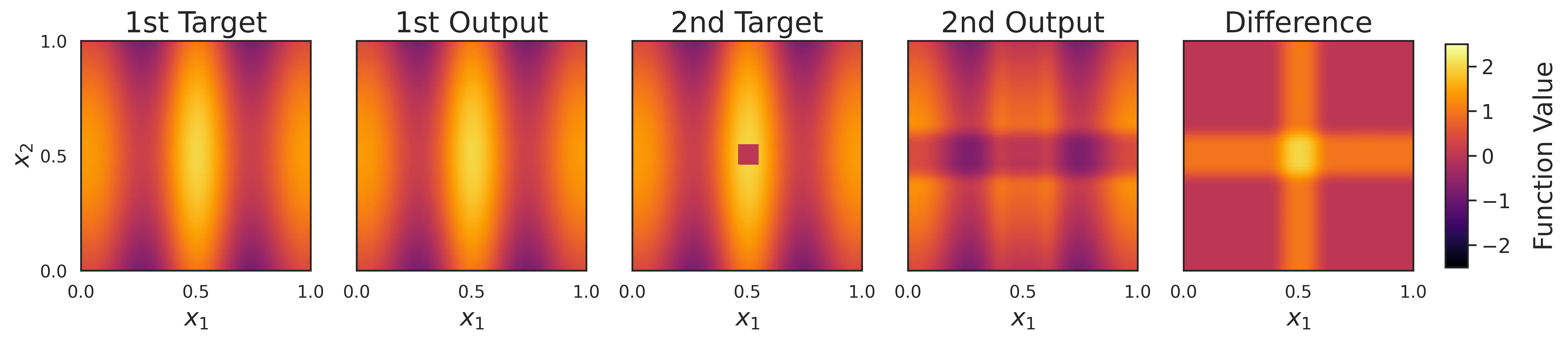}
         \caption{SAM}
         \label{fig:y equals x}
     \end{subfigure}
     \hfill
     \begin{subfigure}[!h]{0.9\textwidth}
         \centering
         \includegraphics[width=\textwidth]{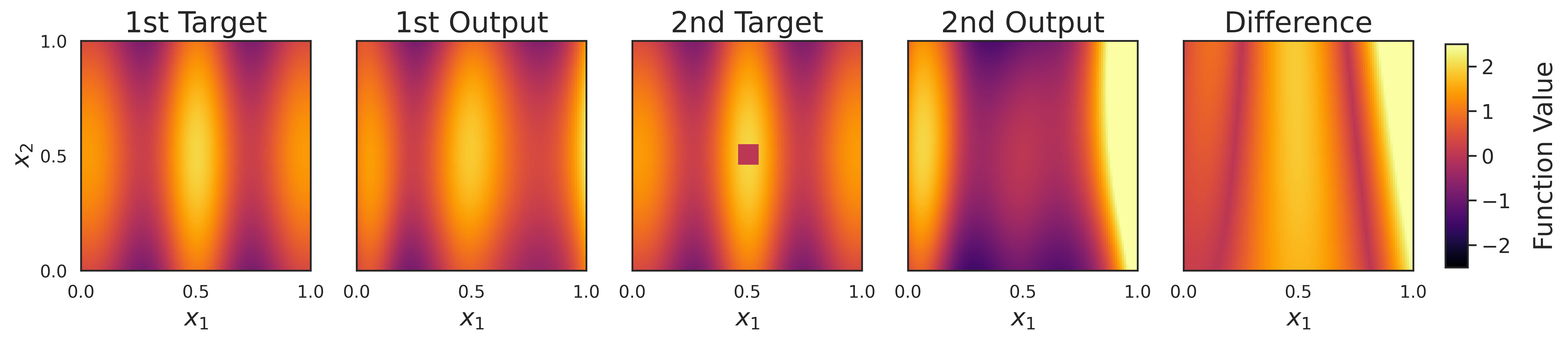}
         \caption{ANN}
         \label{fig:three sin x}
     \end{subfigure}
     \centering
     \begin{subfigure}[!h]{0.9\textwidth}
         \centering
         \includegraphics[width=\textwidth]{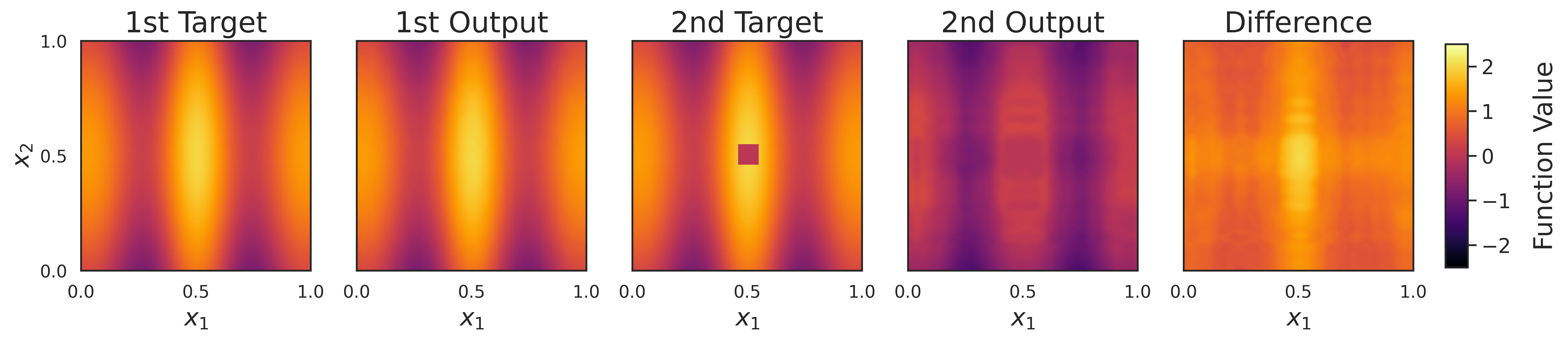}
         \caption{KASAM}
         \label{fig:y equals x}
     \end{subfigure}
     \hfill
     \centering
     \begin{subfigure}[!h]{0.9\textwidth}
         \centering
         \includegraphics[width=\textwidth]{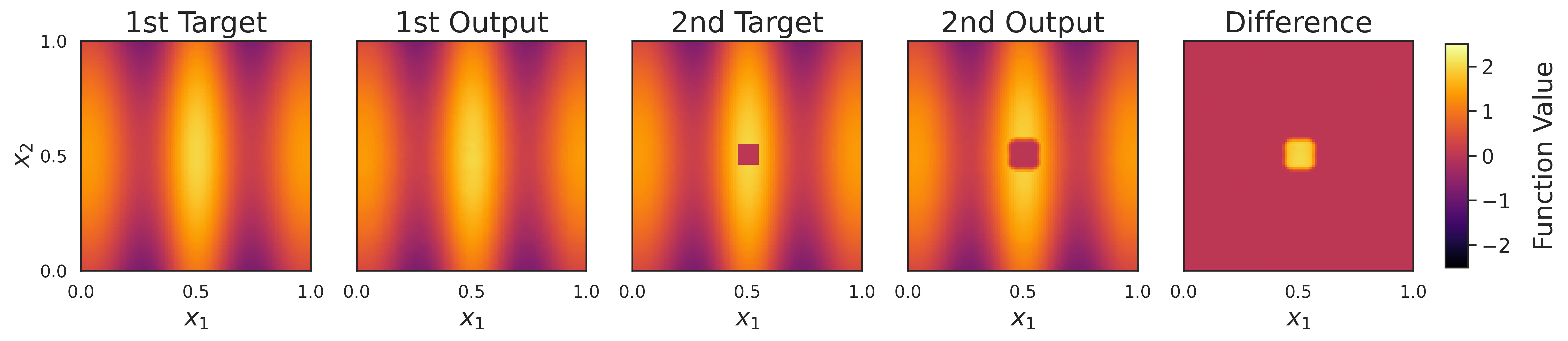}
         \caption{KASAM+PR}
         \label{fig:y equals x}
     \end{subfigure}
     \hfill
        \caption{\small 
Experiment A: outputs of the models and the target functions. Each row corresponds to a different model from top to bottom: SAM, ANN, KASAM, and KASAM+PR. The images for each model from left to right are: Task 1 target function; model output after training on Task 1; Task 2 target function; model output after training on Task 2; The absolute difference between the models' first and second output. A model with nearly perfect memory retention would only differ in the small square region $\left[0.45,0.55 \right] \times \left[0.45,0.55 \right]$ in the centre of the unit square.}
        \label{fig:visualisation_experiment_A}
\end{figure}

The outputs of the models and the target functions were visualised in Figure~\ref{fig:visualisation_experiment_A} with grid-sampled points. Each row corresponds to a different model from top to bottom: SAM, ANN, KASAM, and KASAM+PR. The images for each model from left to right are: Task 1 target function; model output after training on Task 1; Task 2 target function; model output after training on Task 2; The absolute difference between the models' first and second output. A model with nearly perfect memory retention would only differ in the small square region $\left[0.45,0.55 \right] \times \left[0.45,0.55 \right]$ in the centre of the unit square.

For Task 1 all versions of SAM and KASAM could easily fit the target function as visualised in Figure~\ref{fig:visualisation_experiment_A}. The ANN model struggled to fit to the first target function as visualised in Figure~\ref{fig:visualisation_experiment_A}. 

In Task 2 SAM exhibited intrinsic memory retention, except within a cruciform region of overlap between Task 1 and Task 2 as visualised in Figure~\ref{fig:visualisation_experiment_A}, consistent with the developed theory. The ANN suffered catastrophic forgetting while training to output zero in the central region, which fits the training data very well, but it ruined global memory retention. KASAM did not exhibit perfect memory retention on its own as seen in Figure~\ref{fig:visualisation_experiment_A}. KASAM+PR which used pseudo-rehearsal yielded the best memory retention of all four models and boasted nearly perfect performance on the second target function, as seen in Figure~\ref{fig:visualisation_experiment_A}.

%%%%%%%%%%%%%%%%%%%%%%%%%%%%%%%%%%%%%%%%%%%%%%%%%%%%%%%%%%%%%%%%%%%%%%%%%%%%%%%%%%%%%%%%%
\subsection{Experiment B}
%%%%%%%%%%%%%%%%%%%%%%%%%%%%%%%%%%%%%%%%%%%%%%%%%%%%%%%%%%%%%%%%%%%%%%%%%%%%%%%%%%%%%%%%%

The mean and standard deviation of the test MAE over thirty independent trials for each model is shown in Table~\ref{table:B_results_averaged}. The null hypothesis for each pair-wise comparison between models is that they have indistinguishable test errors (threshold is $p<0.0001$). The p-values were calculated from raw data. 

\begin{figure} [!h]
     \centering
     \begin{subfigure}[!h]{0.49\textwidth}
         \centering
         \includegraphics[width=\textwidth]{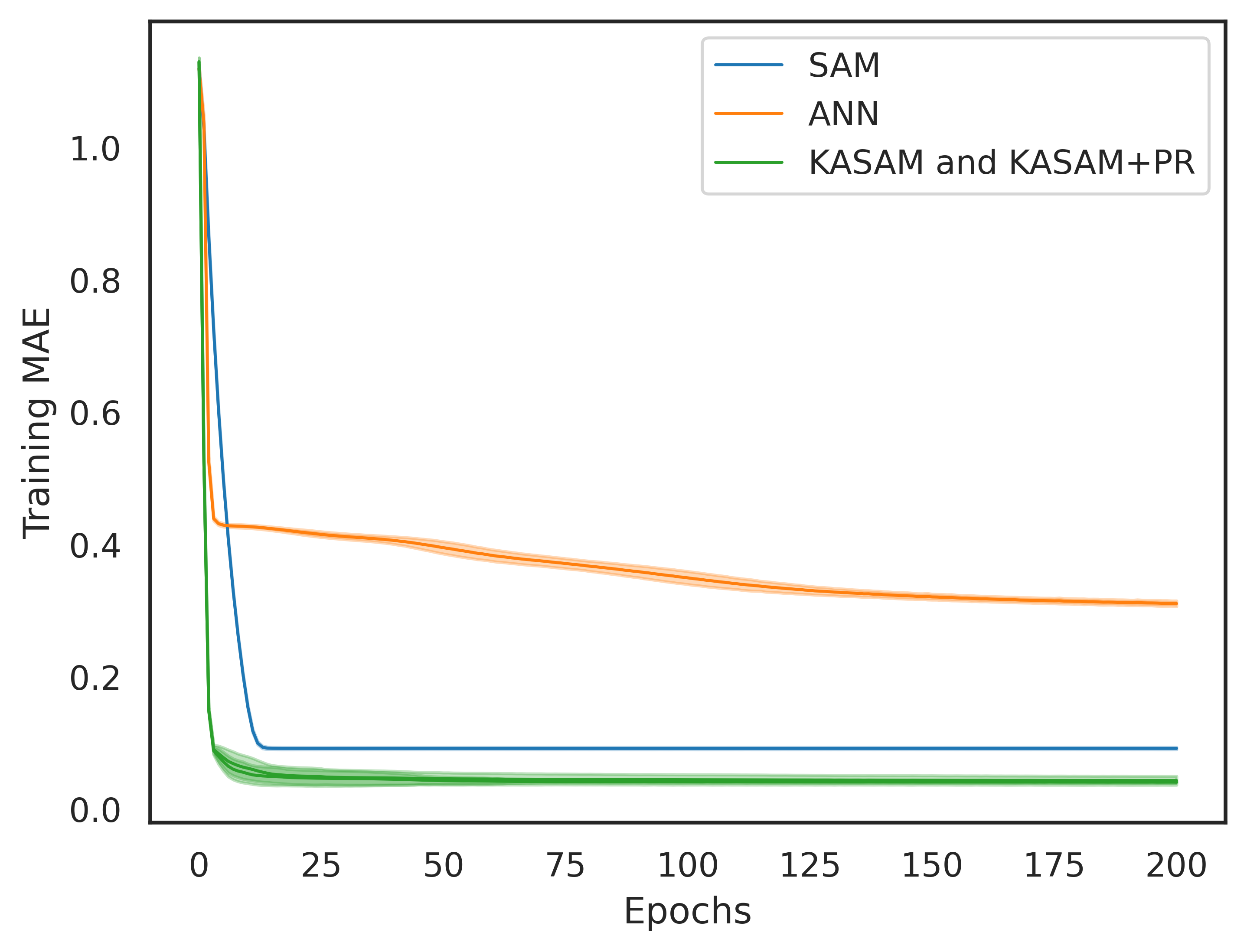}
         \caption{Training loss curve.}
         \label{fig:B_task_1_training_plot}
     \end{subfigure}
     \hfill
     \begin{subfigure}[!h]{0.49\textwidth}
         \centering
         \includegraphics[width=\textwidth]{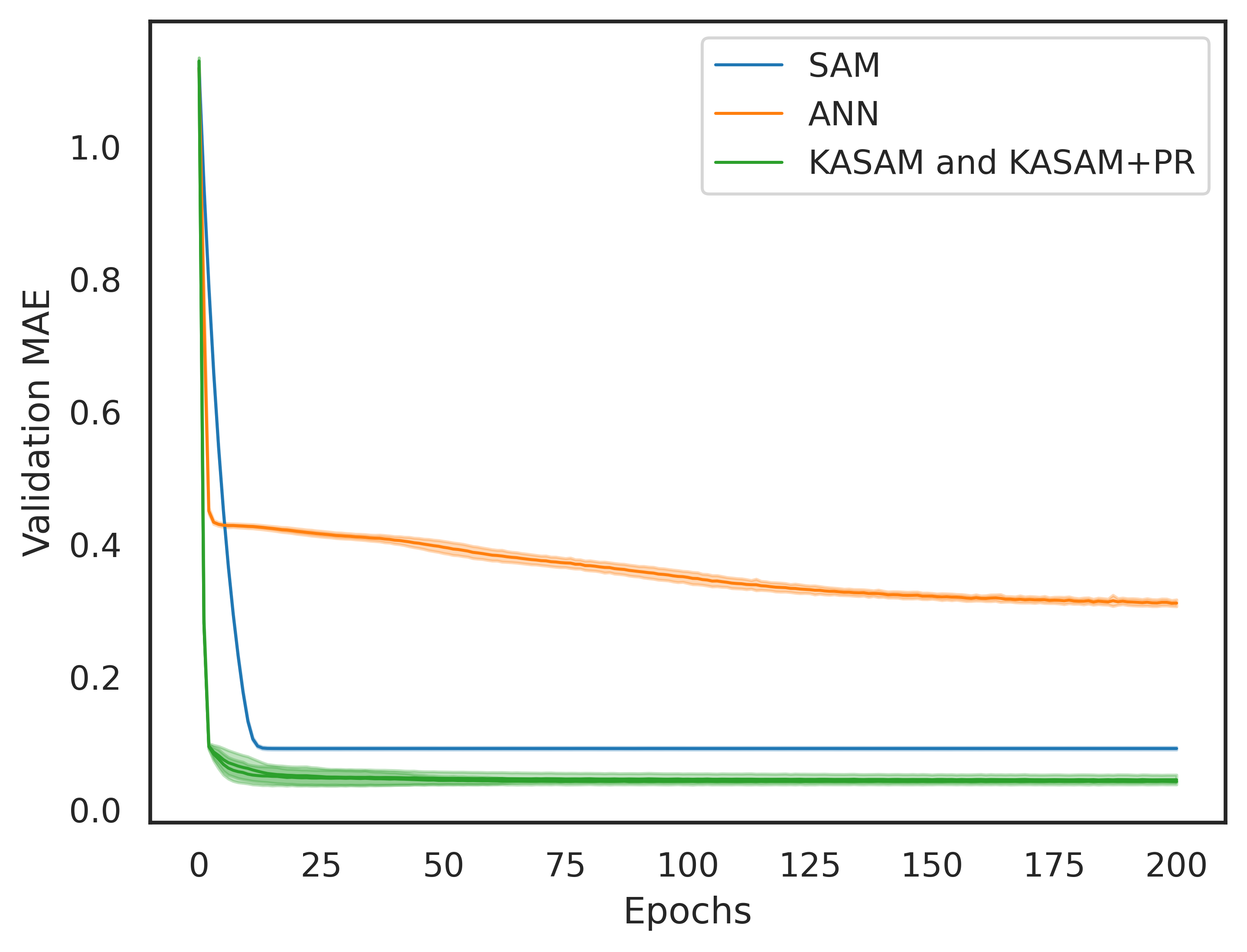}
         \caption{Validation loss curve.}
         \label{fig:B_task_1_validation_plot}
     \end{subfigure}
        \caption{\small 
Experiment B: Task 1 training and validation loss during training. All models were trained on $10000$ training data-points. The initial loss before training is shown at the zeroth epoch.}
        \label{fig:B_task_1_training_validation_plot}
\end{figure}

Task 1 indicated that KASAM and KASAM+PR have indistinguishable test errors, and the null hypothesis was accepted ($p=0.0598$). All other pair-wise comparisons for Task 1 indicated distinguishable test errors, and the null hypothesis was rejected ($p<0.0001$). Rounding the test MSE in Task 1 to a few decimal places shows that KASAM and KASAM+PR have the same test error as shown in Table~\ref{table:B_results_averaged}. SAM had slightly worse performance compared to KASAM. The ANN model was the worst-performing. The training and validation MAE during training is shown in Figure~\ref{fig:B_task_1_training_validation_plot}. The test sets were used for validation as well. The KASAM and KASAM+PR models easily learned the target function, and reasonably quickly. The SAM model couldn't represent the Task 1 target function. The ANN model struggled to learn the Task 1 target function for experiment B as seen in Figure~\ref{fig:B_task_1_training_validation_plot}. 

\begin{figure} [!h]
     \centering
     \begin{subfigure}[!h]{0.49\textwidth}
         \centering
         \includegraphics[width=\textwidth]{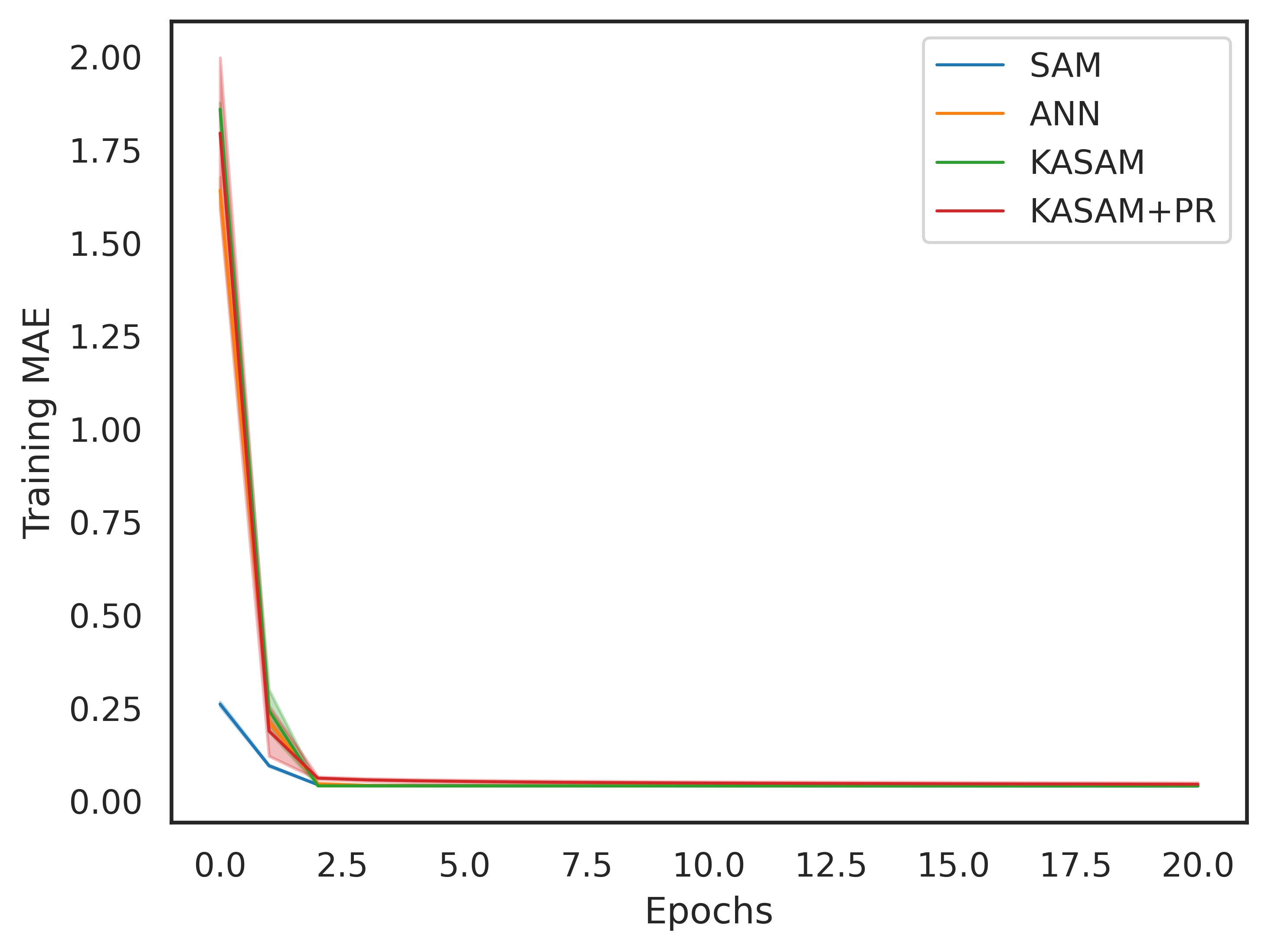}
         \caption{Training loss curve.}
         \label{fig:B_task_2_training_plot}
     \end{subfigure}
     \hfill
     \begin{subfigure}[!h]{0.49\textwidth}
         \centering
         \includegraphics[width=\textwidth]{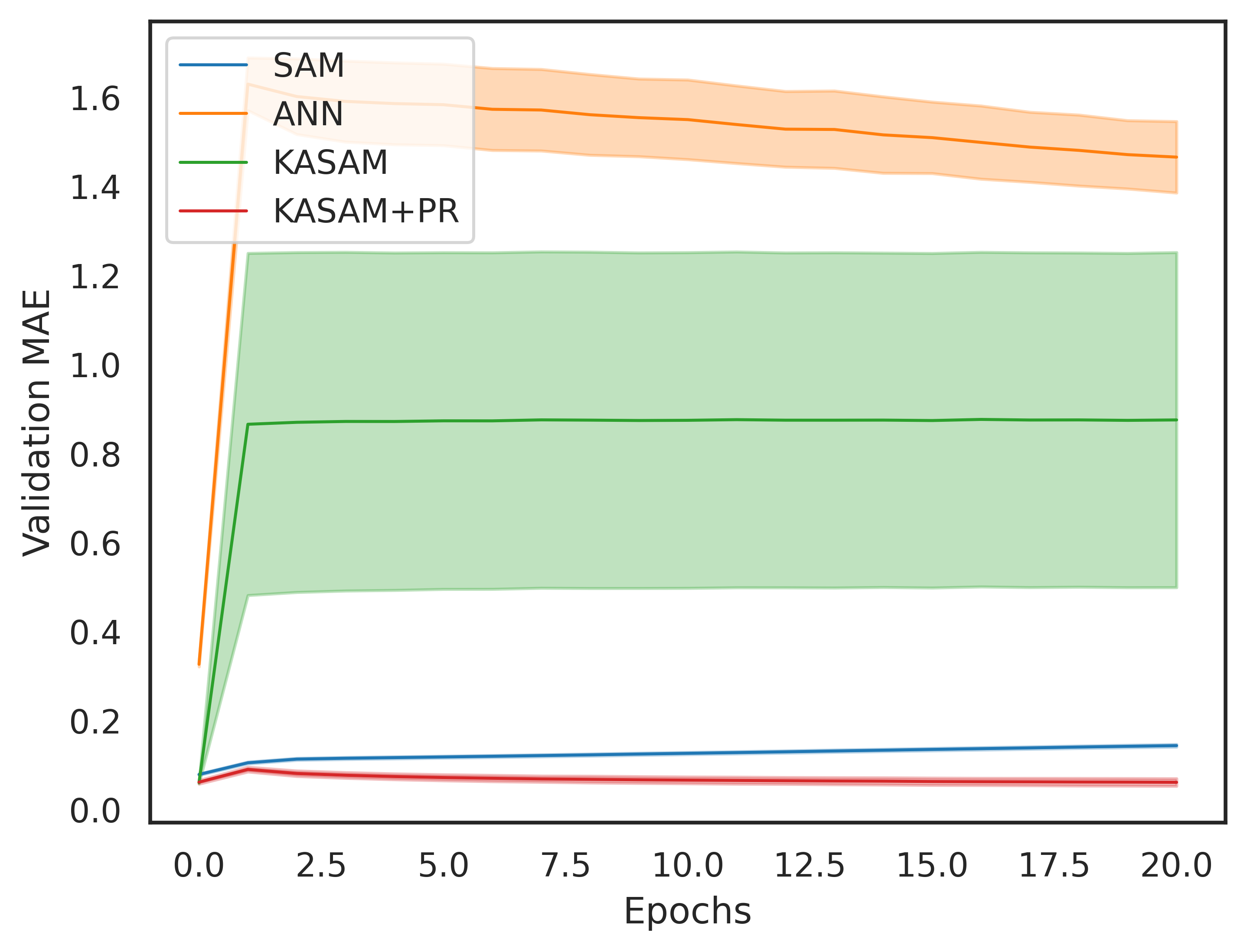}
         \caption{Validation loss curve.}
         \label{fig:B_task_2_validation_plot}
     \end{subfigure}
        \caption{\small 
Experiment B: Task 2 training and validation loss during training. All models were trained on $10000$ training data-points. The initial loss before training is shown at the zeroth epoch.}
        \label{fig:B_task_2_training_validation_plot}
\end{figure}

Task 2 indicated that all four models had distinguishable test errors, and the null hypothesis was rejected ($p<0.0001$) for each pair-wise comparison. KASAM+PR had the best test MAE indicating the benefit of using pseudo-rehearsal techniques with KASAM as shown in Table~\ref{table:B_results_averaged}. The SAM model had the second best performance on Task 2 with some memory retention. The KASAM model alone had the third best performance on Task 2, indicating marginal memory retention. The ANN model had the worst performance and suffered catastrophic forgetting that severely impedes its performance compared to the other models in Table~\ref{table:B_results_averaged}. The training and validation MAE during training is shown in Figure~\ref{fig:B_task_2_training_validation_plot}. The test sets were used for validation as well. All models had similar training loss curves in Figure~\ref{fig:B_task_2_training_validation_plot}. The validation loss curves displayed peculiar dynamics during training Figure~\ref{fig:B_task_2_training_validation_plot}. KASAM+PR performed the best of all models, and pseudo-rehearsal limited catastrophic forgetting and allowed the model to improve after initially degrading in performance. The SAM model degraded in performance and plateaued with little variance. The KASAM and ANN models had the worst performance with a lot of variance in validation MAE, as shown in Figure~\ref{fig:B_task_2_training_validation_plot}. 

\begin{table}[!h]
\centering
\begin{tabular}{|c c c|} 
 \hline
                    & Task 1 MAE                & Task 2 MAE  \\ [0.5ex] 
 \hline\hline
 SAM                & 0.092 (0.002)           & 0.144 (0.003)          \\ 
 ANN                & 0.311 (0.005)           & 1.465 (0.080)          \\ 
 KASAM              & \bf{0.042 (0.001)}      & 0.875 (0.376)          \\ 
 KASAM+PR           & \bf{0.045 (0.007)}      & \bf{0.061 (0.008)}     \\ [0.5ex] 
 \hline
\end{tabular}
\caption{Experiment B: final test mean absolute error (MAE) for Task 1 and Task 2 averaged over 30 trials, rounded to two decimal places.}
\label{table:B_results_averaged}
\end{table}

%%%%%%%%%%%%%%%%%%%%%%%%%%%%%%%%%%%%%%%%%%%%%%%%%%%%%%%%%%%%%%%%%%%%%%%%%%%%%%%%%%%%%%%%%

\begin{figure} [!h]
     \centering
     \begin{subfigure}[!h]{0.9\textwidth}
         \centering
         \includegraphics[width=\textwidth]{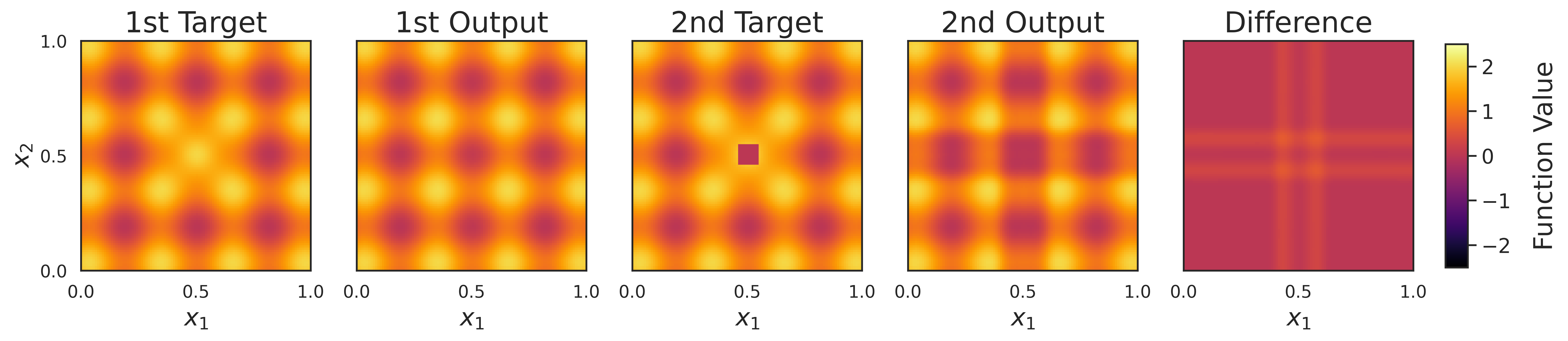}
         \caption{SAM}
         \label{fig:y equals x}
     \end{subfigure}
     \hfill
     \begin{subfigure}[!h]{0.9\textwidth}
         \centering
         \includegraphics[width=\textwidth]{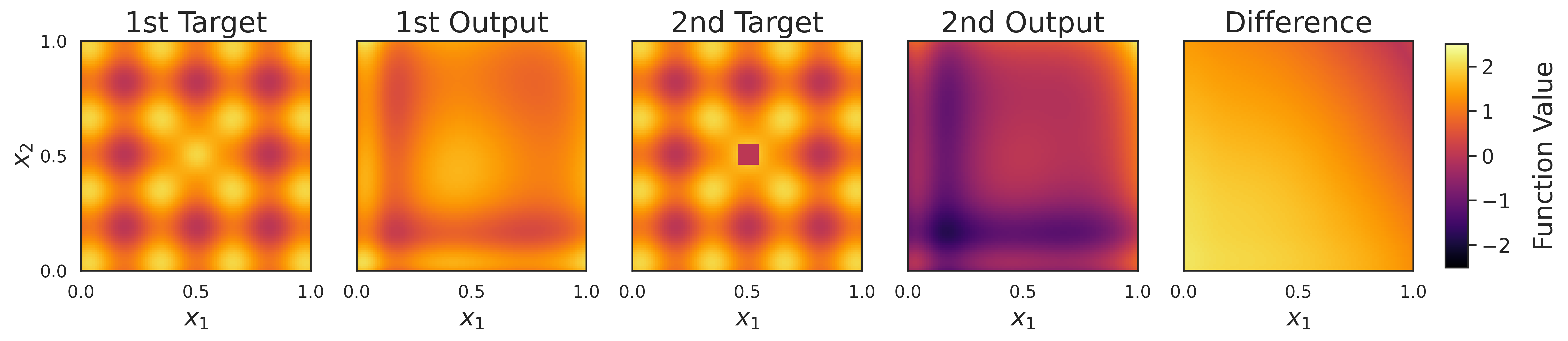}
         \caption{ANN}
         \label{fig:three sin x}
     \end{subfigure}
     \centering
     \begin{subfigure}[!h]{0.9\textwidth}
         \centering
         \includegraphics[width=\textwidth]{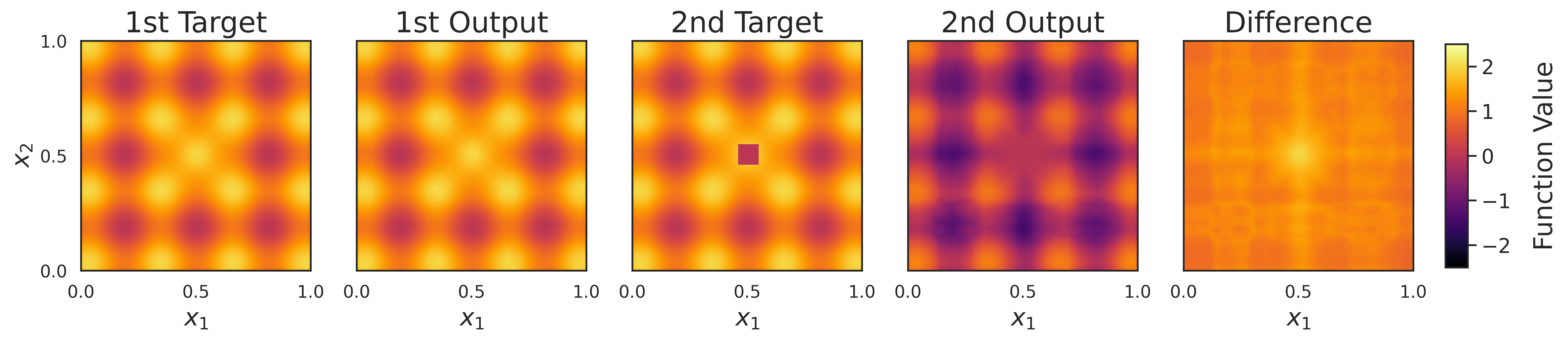}
         \caption{KASAM}
         \label{fig:y equals x}
     \end{subfigure}
     \hfill
     \centering
     \begin{subfigure}[!h]{0.9\textwidth}
         \centering
         \includegraphics[width=\textwidth]{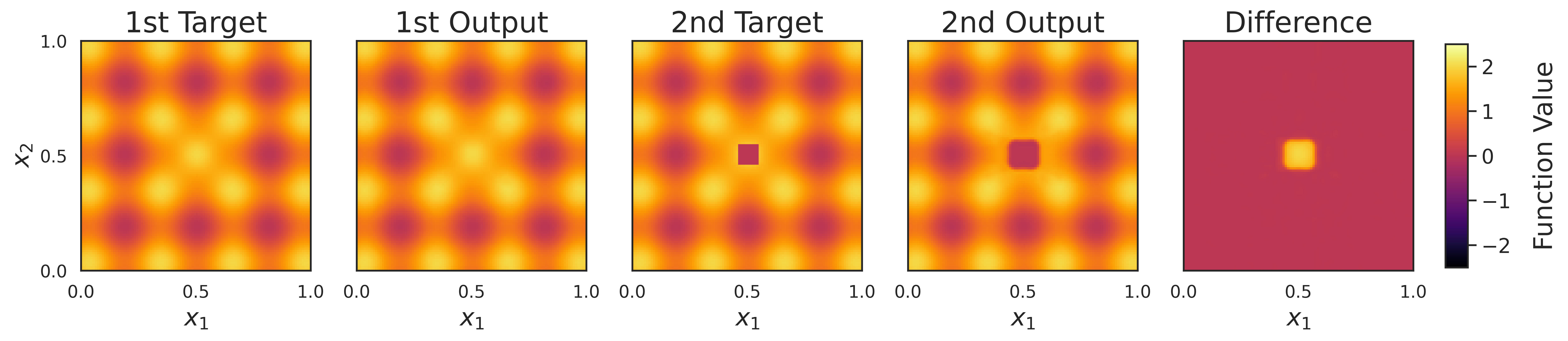}
         \caption{KASAM+PR}
         \label{fig:y equals x}
     \end{subfigure}
     \hfill
        \caption{\small 
Experiment B: outputs of the models and the target functions. Each row corresponds to a different model from top to bottom: SAM, ANN, KASAM, and KASAM+PR. The images for each model from left to right are: Task 1 target function; model output after training on Task 1; Task 2 target function; model output after training on Task 2; The absolute difference between the models' first and second output. A model with nearly perfect memory retention would only differ in the small square region $\left[0.45,0.55 \right] \times \left[0.45,0.55 \right]$ in the centre of the unit square.}
        \label{fig:visualisation_experiment_B}
\end{figure}

The outputs of the models and the target functions were visualised in Figure~\ref{fig:visualisation_experiment_B} with grid-sampled points. Each row corresponds to a different model from top to bottom: SAM, ANN, KASAM, and KASAM+PR. The images for each model from left to right are: Task 1 target function; model output after training on Task 1; Task 2 target function; model output after training on Task 2; The absolute difference between the models' first and second output. A model with nearly perfect memory retention would only differ in the small square region $\left[0.45,0.55 \right] \times \left[0.45,0.55 \right]$ in the centre of the unit square.

For Task 1 KASAM could easily fit the target function as visualised in Figure~\ref{fig:visualisation_experiment_B}. The SAM and ANN model struggled to fit to the first target function as visualised in Figure~\ref{fig:visualisation_experiment_B}. 

In Task 2 SAM exhibited intrinsic memory retention, except within a cruciform region of overlap between Task 1 and Task 2 as visualised in Figure~\ref{fig:visualisation_experiment_B}, consistent with the developed theory. The ANN suffered catastrophic forgetting while training to output zero in the central region, which fits the training data very well, but it ruined global memory retention. KASAM did not exhibit perfect memory retention on its own as seen in Figure~\ref{fig:visualisation_experiment_B}. KASAM+PR which used pseudo-rehearsal yielded the best memory retention of all four models and boasted nearly perfect performance on the second target function, as seen in Figure~\ref{fig:visualisation_experiment_B}.

%%%%%%%%%%%%%%%%%%%%%%%%%%%%%%%%%%%%%%%%%%%%%%%%%%%%%%%%%%%%%%%%%%%%%%%%%%%%%%%%%%%%%%%%%

\subsection{Experiment C}

The mean and standard deviation of the test MAE over thirty independent trials for each model is shown in Table~\ref{table:C_results_averaged}. The null hypothesis for each pair-wise comparison between models is that they have indistinguishable test errors (threshold is $p<0.0001$). The p-values were calculated from raw data. 

\begin{figure} [!h]
     \centering
     \begin{subfigure}[!h]{0.49\textwidth}
         \centering
         \includegraphics[width=\textwidth]{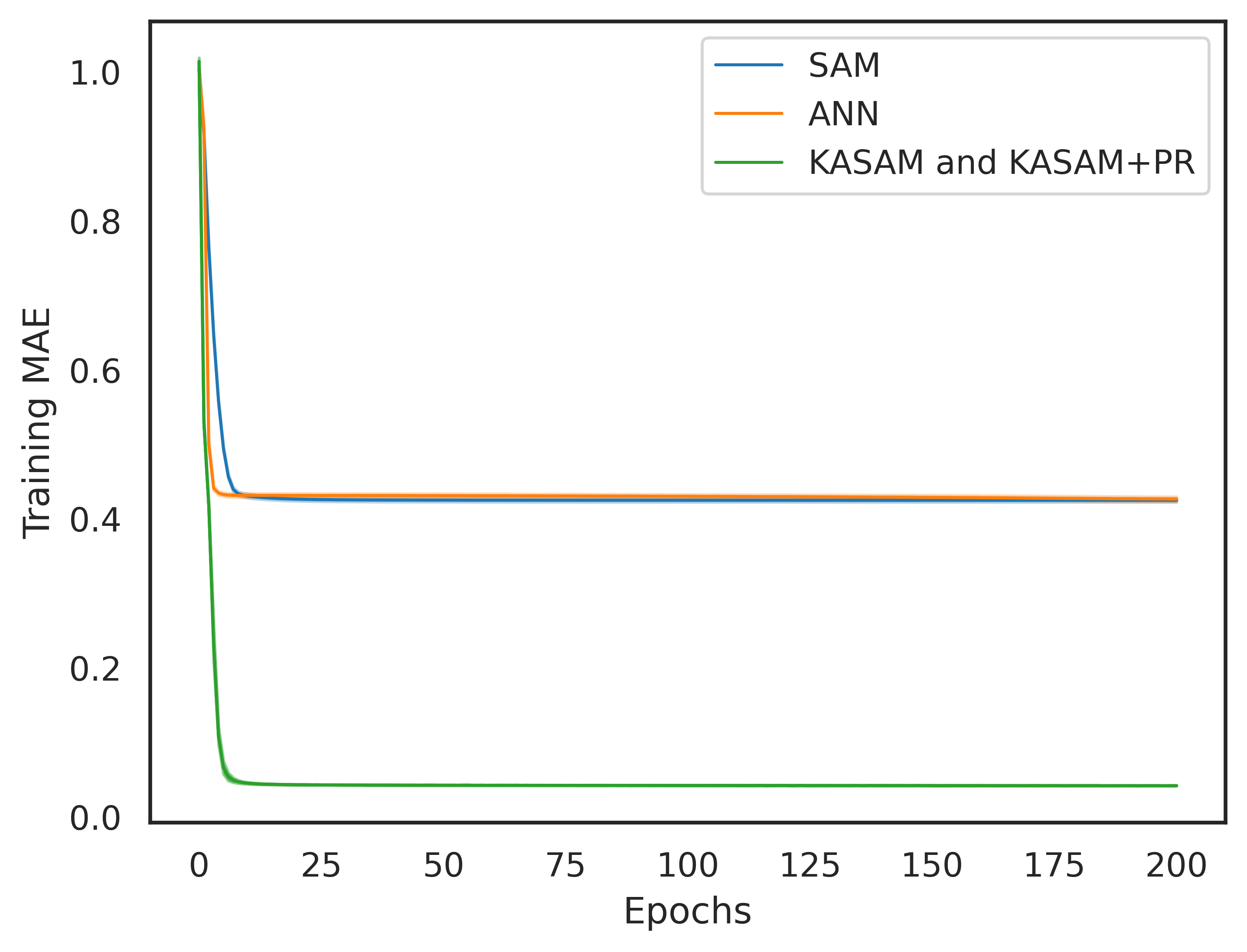}
         \caption{Training loss curve.}
         \label{fig:C_task_1_training_plot}
     \end{subfigure}
     \hfill
     \begin{subfigure}[!h]{0.49\textwidth}
         \centering
         \includegraphics[width=\textwidth]{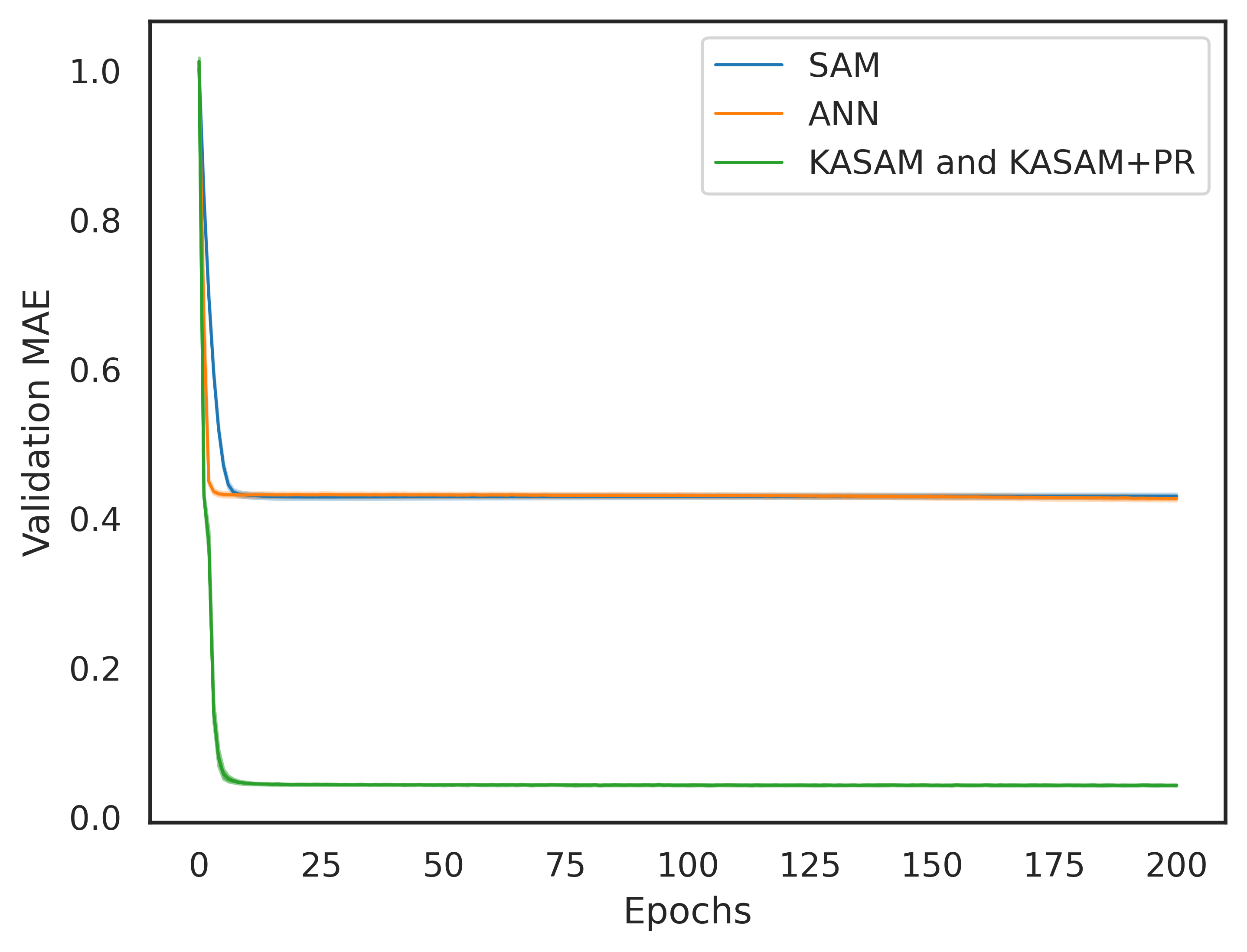}
         \caption{Validation loss curve.}
         \label{fig:C_task_1_validation_plot}
     \end{subfigure}
        \caption{\small 
Experiment C: Task 1 training and validation loss during training. All models were trained on $10000$ training data-points. The initial loss before training is shown at the zeroth epoch.}
        \label{fig:C_task_1_training_validation_plot}
\end{figure}

Task 1 test MAE indicated that KASAM and KASAM+PR have indistinguishable test errors, and the null hypothesis was accepted ($p=0.6018$). KASAM and KASAM had the best performance. The SAM and ANN models also have the same test error ($p=0.0007$), and the worst performance. All other pair-wise comparisons for Task 1 indicated distinguishable test errors, and the null hypothesis was rejected ($p<0.0001$). The test MSE in Task 1 is rounded to a few decimal places and presented in Table~\ref{table:C_results_averaged}. The training and validation MAE during training is shown in Figure~\ref{fig:C_task_1_training_validation_plot}. The test sets were used for validation as well. The KASAM and KASAM+PR models easily learned the target function, and reasonably quickly. The SAM and ANN model struggled to learn the Task 1 target function for experiment C as seen in Figure~\ref{fig:C_task_1_training_validation_plot}. 

\begin{figure} [!h]
     \centering
     \begin{subfigure}[!h]{0.49\textwidth}
         \centering
         \includegraphics[width=\textwidth]{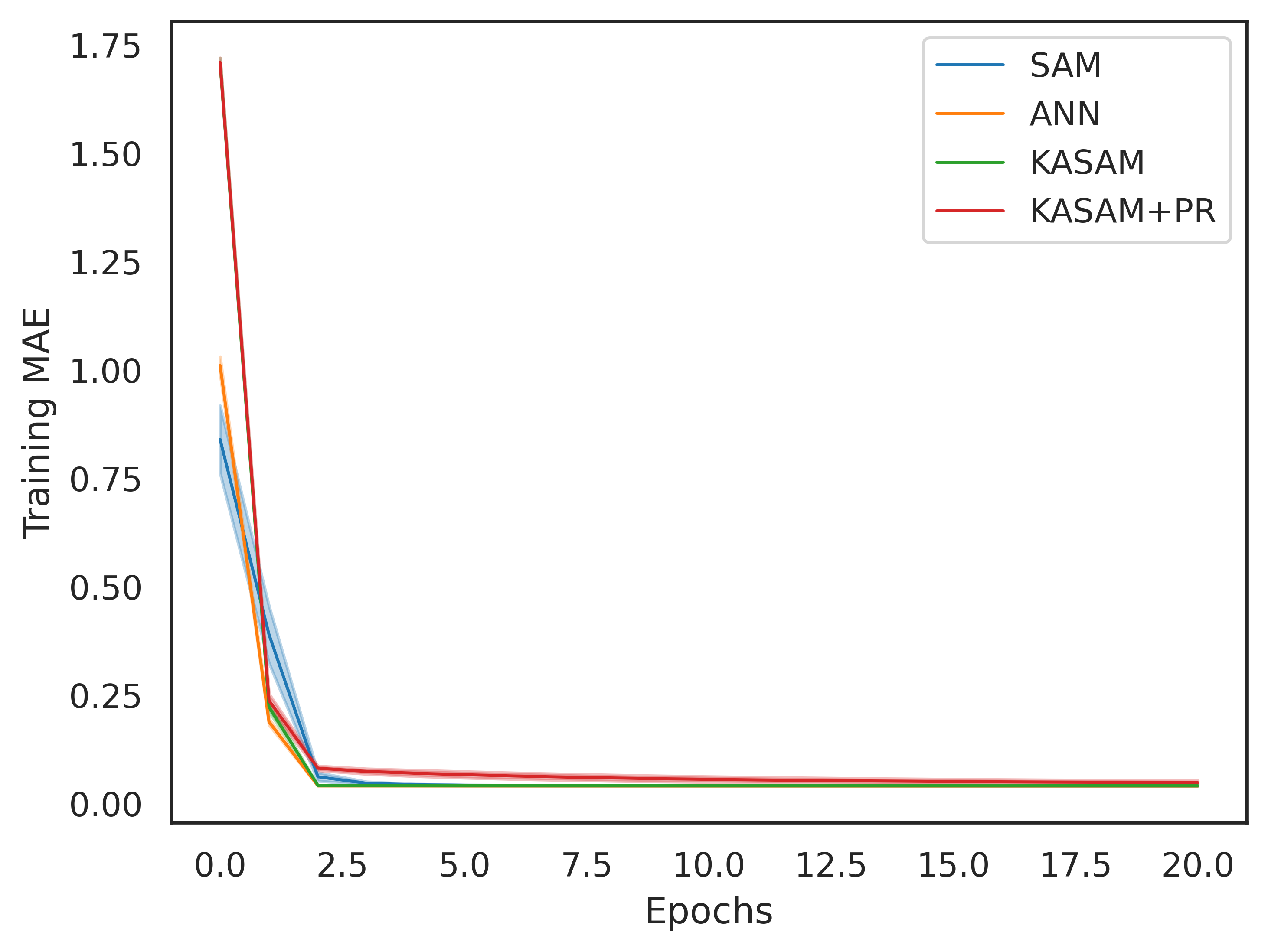}
         \caption{Training loss curve.}
         \label{fig:C_task_2_training_plot}
     \end{subfigure}
     \hfill
     \begin{subfigure}[!h]{0.49\textwidth}
         \centering
         \includegraphics[width=\textwidth]{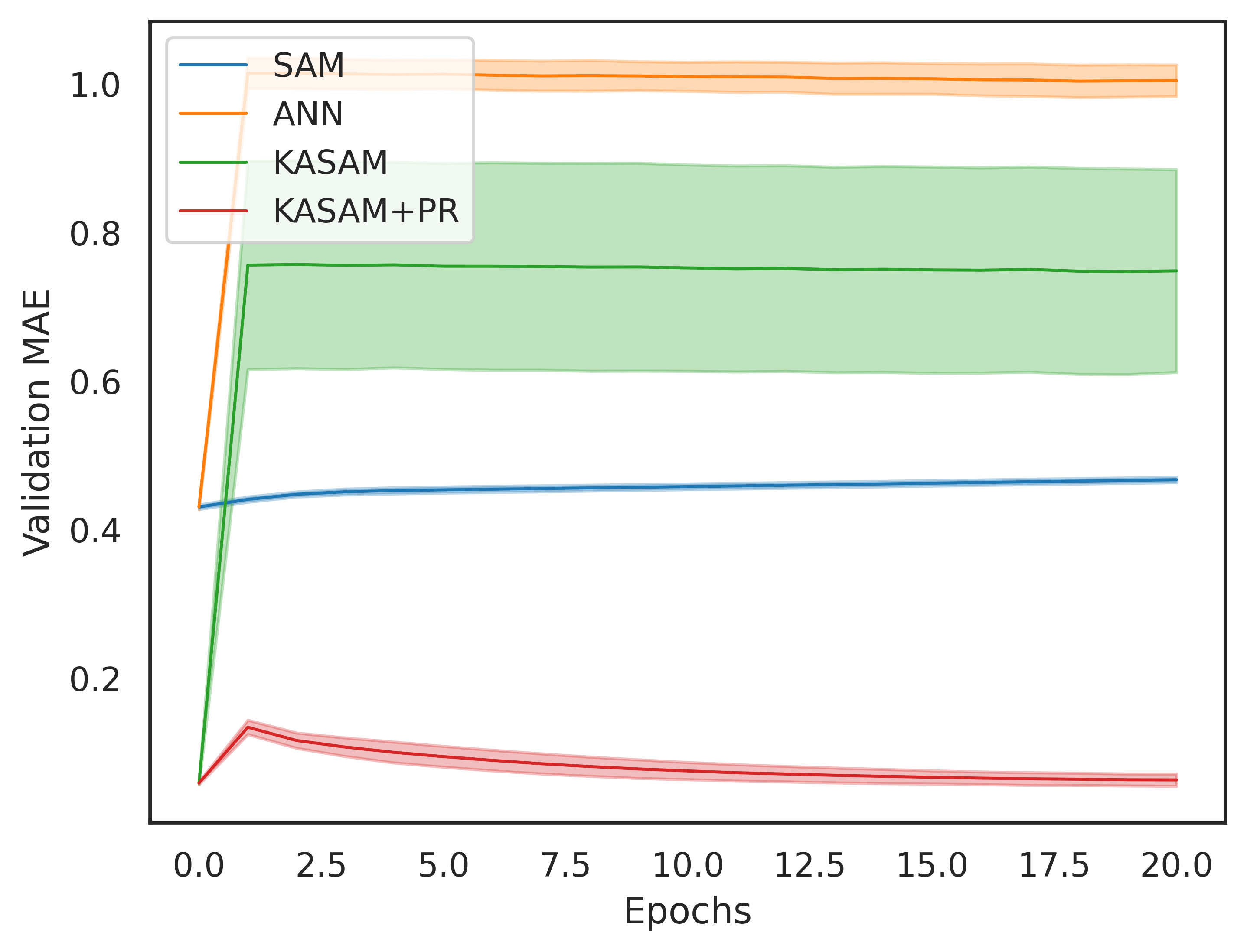}
         \caption{Validation loss curve.}
         \label{fig:C_task_2_validation_plot}
     \end{subfigure}
        \caption{\small 
Experiment C: Task 2 training and validation loss during training. All models were trained on $10000$ training data-points. The initial loss before training is shown at the zeroth epoch.}
        \label{fig:C_task_2_training_validation_plot}
\end{figure}

Task 2 indicated that all four models had distinguishable test errors, and the null hypothesis was rejected ($p<0.0001$) for each pair-wise comparison. KASAM+PR had the best test MAE indicating the benefit of using pseudo-rehearsal techniques with KASAM as shown in Table~\ref{table:C_results_averaged}. The SAM model had the second best performance on Task 2 with some memory retention. The KASAM model alone had the third best performance on Task 2, indicating marginal memory retention. The ANN model had the worst performance and suffered catastrophic forgetting that severely impedes its performance compared to the other models in Table~\ref{table:C_results_averaged}. The training and validation MAE during training is shown in Figure~\ref{fig:C_task_2_training_validation_plot}. The test sets were used for validation as well. All models had similar training loss curves in Figure~\ref{fig:C_task_2_training_validation_plot}. The validation loss curves displayed interesting dynamics during training Figure~\ref{fig:C_task_2_training_validation_plot}. KASAM+PR performed the best of all models, and pseudo-rehearsal limited catastrophic forgetting and allowed the model to improve after initially degrading in performance. The SAM model degraded in performance and plateaued with little variance. The KASAM and ANN models had the worst performance with a lot of variance in validation MAE, as shown in Figure~\ref{fig:C_task_2_training_validation_plot}. 

\begin{table}[!h]
\centering
\begin{tabular}{|c c c|} 
 \hline
                    & Task 1 MAE            & Task 2 MAE            \\ [0.5ex] 
 \hline\hline
 SAM                & 0.430 (0.003)           & 0.467 (0.004)          \\ 
 ANN                & 0.427 (0.003)           & 1.004 (0.021)          \\ 
 KASAM              & \bf{0.042 (0.001)}      & 0.748 (0.137)          \\ 
 KASAM+PR           & \bf{0.042 (0.001)}      & \bf{0.063 (0.008)}     \\ [0.5ex] 
 \hline
\end{tabular}
\caption{Experiment C: final test mean absolute error (MAE) for Task 1 and Task 2 averaged over 30 trials, rounded to two decimal places.}
\label{table:C_results_averaged}
\end{table}

%%%%%%%%%%%%%%%%%%%%%%%%%%%%%%%%%%%%%%%%%%%%%%%%%%%%%%%%%%%%%%%%%%%%%%%%%%%%%%%%%%%%%%%%%

\begin{figure} [!h]
     \centering
     \begin{subfigure}[!h]{0.9\textwidth}
         \centering
         \includegraphics[width=\textwidth]{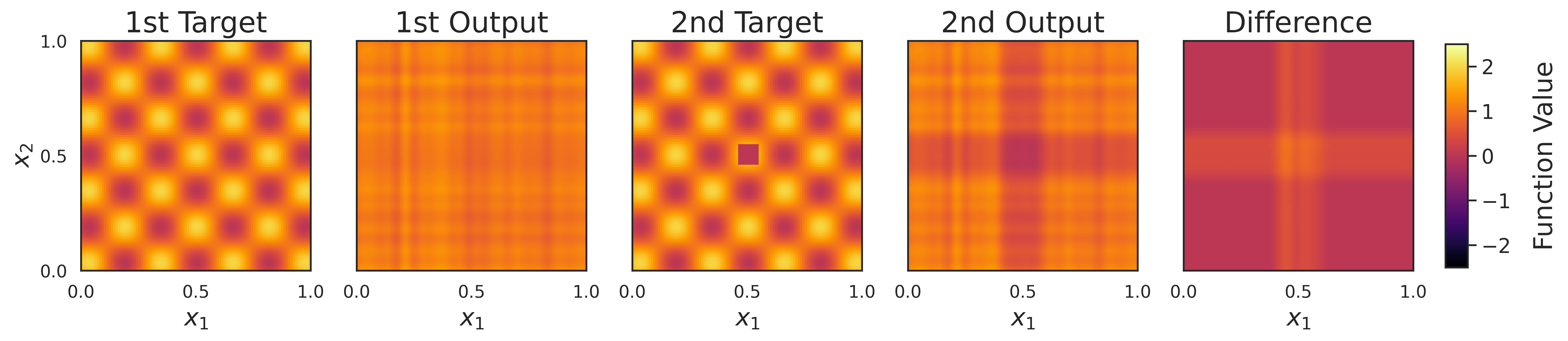}
         \caption{SAM}
         \label{fig:y equals x}
     \end{subfigure}
     \hfill
     \begin{subfigure}[!h]{0.9\textwidth}
         \centering
         \includegraphics[width=\textwidth]{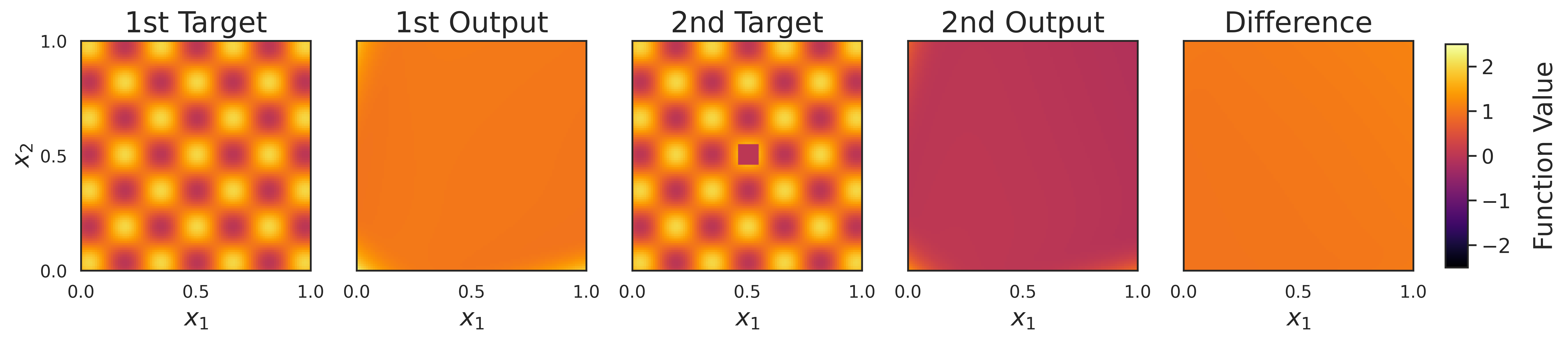}
         \caption{ANN}
         \label{fig:three sin x}
     \end{subfigure}
     \centering
     \begin{subfigure}[!h]{0.9\textwidth}
         \centering
         \includegraphics[width=\textwidth]{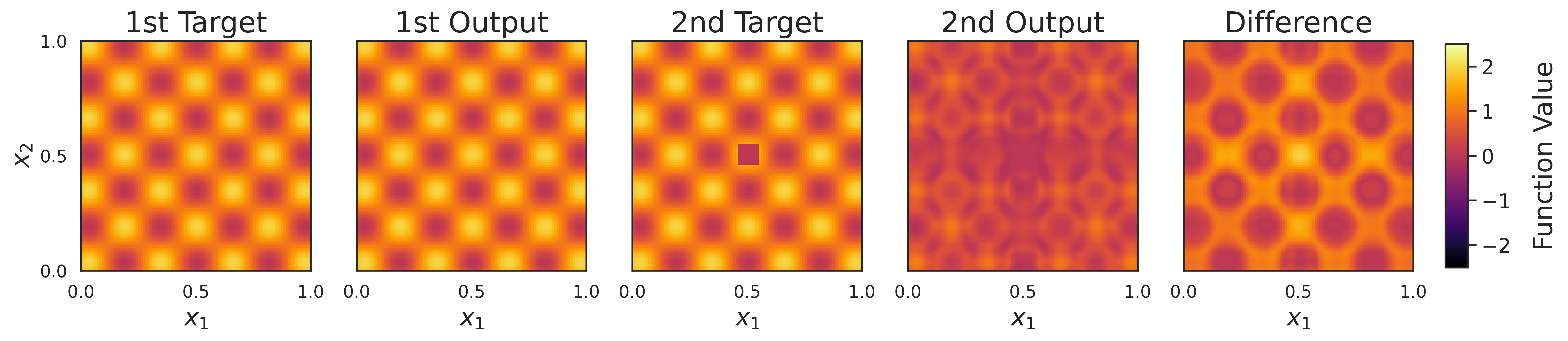}
         \caption{KASAM}
         \label{fig:y equals x}
     \end{subfigure}
     \hfill
     \centering
     \begin{subfigure}[!h]{0.9\textwidth}
         \centering
         \includegraphics[width=\textwidth]{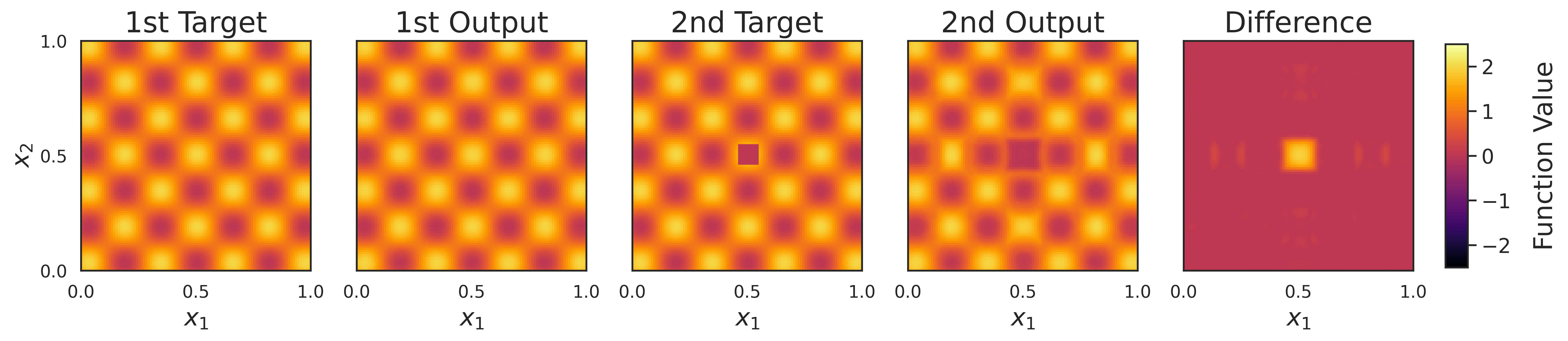}
         \caption{KASAM+PR}
         \label{fig:y equals x}
     \end{subfigure}
     \hfill
        \caption{\small 
Experiment C: outputs of the models and the target functions. Each row corresponds to a different model from top to bottom: SAM, ANN, KASAM, and KASAM+PR. The images for each model from left to right are: Task 1 target function; model output after training on Task 1; Task 2 target function; model output after training on Task 2; The absolute difference between the models' first and second output. A model with nearly perfect memory retention would only differ in the small square region $\left[0.45,0.55 \right] \times \left[0.45,0.55 \right]$ in the centre of the unit square.}
        \label{fig:visualisation_experiment_C}
\end{figure}

The outputs of the models and the target functions were visualised in Figure~\ref{fig:visualisation_experiment_C} with grid-sampled points. Each row corresponds to a different model from top to bottom: SAM, ANN, KASAM, and KASAM+PR. The images for each model from left to right are: Task 1 target function; model output after training on Task 1; Task 2 target function; model output after training on Task 2; The absolute difference between the models' first and second output. A model with nearly perfect memory retention would only differ in the small square region $\left[0.45,0.55 \right] \times \left[0.45,0.55 \right]$ in the centre of the unit square.

For Task 1 all versions of KASAM could easily fit the target function as visualised in Figure~\ref{fig:visualisation_experiment_C}. The SAM and ANN model struggled to fit to the first target function as visualised in Figure~\ref{fig:visualisation_experiment_C}. 

In Task 2 SAM exhibited intrinsic memory retention, except within a cruciform region of overlap between Task 1 and Task 2 as visualised in Figure~\ref{fig:visualisation_experiment_C}, consistent with the developed theory. The ANN suffered catastrophic forgetting while training to output zero in the central region, which fits the training data very well, but it ruined global memory retention. KASAM did not exhibit perfect memory retention on its own as seen in Figure~\ref{fig:visualisation_experiment_C}. KASAM+PR which used pseudo-rehearsal yielded the best memory retention of all four models and boasted nearly perfect performance on the second target function, as seen in Figure~\ref{fig:visualisation_experiment_C}.

%%%%%%%%%%%%%%%%%%%%%%%%%%%%%%%%%%%%%%%%%%%%%%%%%%%%%%%%%%%%%%%%%%%%%%%%%%%%%%%%%%%%%%%%%
\section{Conclusions}\label{sec:conclusions}
%%%%%%%%%%%%%%%%%%%%%%%%%%%%%%%%%%%%%%%%%%%%%%%%%%%%%%%%%%%%%%%%%%%%%%%%%%%%%%%%%%%%%%%%%
%theory, technical, and empirical.

%The paper's main contribution is developing theoretical models to combat catastrophic forgetting, implemented the approaches in existing software, and demonstrating that the developed models work.

%The paper contributes with theory, writing software, and demonstrating that it works with a simple problem

This paper contributes in three ways: theoretically, technically, and empirically. Catastrophic forgetting was analysed and theoretical models, namely SAM and KASAM, were developed to combat catastrophic forgetting. The developed models were implemented in TensorFlow, and released for public use. The models were analysed on a simple problem to empirically demonstrate their effectiveness.

The paper introduced and implemented Spline Additive Models (SAMs) to demonstrate their robustness to catastrophic forgetting, and that SAM itself is not a universal function approximator, but is still useful for many potential applications. The Kolmogorov-Arnold Spline Additive Model (KASAM) was introduced and implemented. KASAM is shown to be a universal function approximator that is expressive, but more susceptible to catastrophic forgetting than SAM. It is unknown if tractable models exist with better guaranteed bounds for catastrophic forgetting.

The empirical scope of the paper was limited to target functions of two variables, mainly for demonstration purposes. The statistical analysis and detailed inspection of the results show that SAM exhibits intrinsic memory retention that is robust to catastrophic forgetting. The memory retention of SAM is not perfect: SAM exhibits cross-shaped regions of overlapping interference in higher dimensional models. SAM is also shown not to be a universal function approximator. The extension of SAM to a universal function approximator leads to a more expressive model: KASAM. KASAM has limited intrinsic memory retention, but is a universal function approximator.

Conventional or typical neural networks based on affine or linear transformations can be susceptible to catastrophic forgetting. A more typical artificial neural network with the exact same structure and activation functions as KASAM, with randomly initialised and trainable parameters, performed significantly worse than KASAM. The feed-forward artificial neural network (ANN) had the capacity to implement KASAM and spline functions, but it did not exploit this potential in any appreciable way during training and evaluation. KASAM exhibited superior performance with some weights being chosen constants, as compared to having all of its parameters being randomly initialised and trainable.

KASAM in combination with other regularisation, data-augmentation and training techniques can mitigate catastrophic forgetting. KASAM in combination with pseudo-rehearsal had the best performance of all models considered for the chosen regression problem. Pseudo-rehearsal works well for memory retention in low-dimensional problems. 

%%%%%%%%%%%%%%%%%%%%%%%%%%%%%%%%%%%%%%%%%%%%%%%%%%%%%%%%%%%%%%%%%%%%%%%%%%%%%%%%%%%%%%%%%
\section{Opportunities for Further Research}\label{sec:opportunities}
%%%%%%%%%%%%%%%%%%%%%%%%%%%%%%%%%%%%%%%%%%%%%%%%%%%%%%%%%%%%%%%%%%%%%%%%%%%%%%%%%%%%%%%%%

This paper explored a few models and ideas to combat catastrophic forgetting. It is an open question if there are other approaches to combat catastrophic forgetting.

Future research can explore: More efficient implementation; Letting the sub-intervals vary during training instead of using uniform partitions; experimenting with different target functions; experimenting with higher dimensional problems; increasing the density of basis functions during training; the incorporation of B-spline functions into other models such as recurrent neural networks, LSTMs, GRUs or reservoir computers; evaluating the bias-variance decomposition for over-parameterised B-spline functions; fitness landscape analysis of B-spline functions; or different initialisations of parameters for KASAM. 

%%%%%%%%%%%%%%%%%%%%%%%%%%%%%%%%%%%%%%%%%%%%%%%%%%%%%%%%%%%%%%%%%%%%%%%%%%%%%%%%%%%%%%%%%
%%%%%%%%%%%%%%%%%%%%%%%%%%%%%%%%%%%%%%%%%%%%%%%%%%%%%%%%%%%%%%%%%%%%%%%%%%%%%%%%%%%%%%%%%
\bibliography{kasam}
%%%%%%%%%%%%%%%%%%%%%%%%%%%%%%%%%%%%%%%%%%%%%%%%%%%%%%%%%%%%%%%%%%%%%%%%%%%%%%%%%%%%%%%%%
\end{document}